\documentclass[10pt,twocolumn,letterpaper]{article}

\usepackage[pagenumbers]{cvpr} % To force page numbers, e.g. for an arXiv version

\makeatletter
\newcommand{\paragraphnew}{\@startsection{paragraph}{4}{\z@}%
  {.5ex \@plus .1ex \@minus .2ex}%
  {-.5em}%
  {\normalfont\normalsize\bfseries}}
\makeatother

\usepackage{subcaption}
\usepackage{booktabs}
\usepackage{graphicx}
\usepackage{xcolor, colortbl}
\usepackage{multirow}
\usepackage{adjustbox}
\usepackage{subcaption}
\usepackage{float,tabularx}

\definecolor{cvprblue}{rgb}{0.21,0.49,0.74}
\usepackage[pagebackref,breaklinks,colorlinks,allcolors=cvprblue]{hyperref}
\def\confName{CVPR}
\def\confYear{2027}

\title{{See the Answer, Learn to Embed}: \\ Learning Universal Multimodal Embeddings from Hindsight}
\author{Yoonki Cho
  \quad Youngju Na
  \quad Taegeun Yang
  \quad Jinhwan Seo
  \quad Sung-Eui Yoon
  \\[0.5ex]
  KAIST
  \\[0.2ex]
  \normalsize{\url{https://yoonkicho.github.io/HEME}}
}

\begin{document}
\maketitle
\begin{abstract}
Universal Multimodal Embedding (UME) models map heterogeneous multimodal inputs into a shared embedding space, where pairwise similarity supports diverse downstream tasks.
Recent reasoning-driven UME approaches use intermediate reasoning to generate additional context for embedding, but often incur extra computation at inference.
In this paper, we explore the matched candidate (\ie, the paired target) as an alternative source of context, asking whether a model can ``see the answer'' during training and ``learn to embed'' without it at inference.
However, direct conditioning on this privileged context can introduce candidate-specific information unavailable from the query alone or cause the candidate-conditioned query representation to collapse toward the candidate embedding.
To address this, we introduce \textit{Hindsight-Enhanced Multimodal Embeddings} (\textbf{HEME}), which learns query-grounded hindsight representations from such candidate context.
HEME first trains a hindsight teacher with a hindsight attention mask, mismatched-candidate conditioning, and query-token alignment to limit excessive reliance on the candidate.
It then freezes the teacher and distills its hindsight representation and similarity distribution into a standalone student embedder.
At inference, the student independently encodes queries and candidates without candidate conditioning or the hindsight teacher.
Experiments on MMEB-V2 and the agent and text tasks of MMEB-V3 demonstrate the effectiveness of HEME and show that it enables more efficient inference than explicit reasoning-driven methods.
\end{abstract}
    
\section{Introduction}
\label{sec:introduction}

\begin{figure}[t]
    \centering
    \includegraphics[width=\columnwidth]{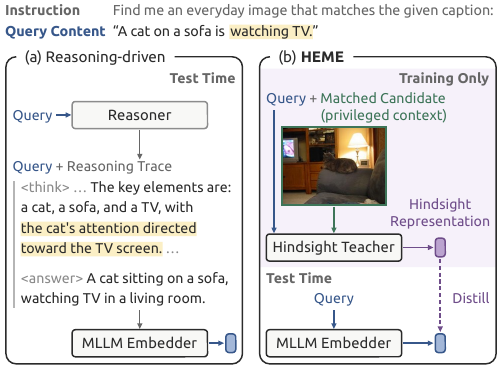}
    \caption{\textbf{Reasoning at test time vs.\ hindsight during training.}
    (a) Reasoning-driven UME approaches generate additional context through intermediate reasoning at test time.
    (b) HEME instead uses the matched candidate as privileged context during training: a hindsight teacher produces a candidate-conditioned hindsight representation, which is distilled into a standalone embedder.
    At test time, HEME encodes queries and candidates independently without the hindsight teacher or additional reasoning steps.}
    \vspace{-3mm}
\label{fig:introduction}
\end{figure}

Universal Multimodal Embedding (UME) models aim to represent a wide range of modalities, such as text, images, videos, and visual documents, as well as their combinations, within a shared embedding space for diverse downstream tasks~\cite{meng2025vlm2vecv2, zhang2025bridging, gu2025breaking}.
Beyond aligning such heterogeneous inputs, a core challenge is to determine which aspects of an input should be emphasized in its embedding for a given task.
Multimodal Large Language Models (MLLMs)~\cite{liu2023visual, wang2024qwen2vl} provide a strong foundation for this goal by interpreting multimodal content in the context of task instructions~\cite{jiang2024vlm2vec, jiang2024e5v, lin2025mmembed}.
Yet even with such instructions, standard embedding objectives supervise which candidate should match a query, but do not directly indicate \emph{which query cues} make that candidate relevant.
Consequently, models can struggle when the match depends on fine-grained attributes or relations that require complex interpretation~\cite{cui2025think, liu2026rematch}.
To address this, recent reasoning-driven UME approaches use intermediate reasoning to generate additional context for embedding, but often require extra computation at inference~\cite{lan2026umer1, wang2026mmembr1, wu2026beyond}.

In this paper, we explore an alternative source of context available during training: the \textit{matched candidate} (\ie, the ground-truth target paired with the query).
Observing this target can help reveal which details of the query are relevant to the match.
For example, in Fig.~\ref{fig:introduction}, given the caption ``A cat on a sofa is watching TV,'' the matched image can help identify the relevant details in the caption, including the relation between the cat and the TV.
We thus allow an MLLM embedder to process the query jointly with its matched candidate during training, while preserving independent query and candidate encoding at inference.
By treating the matched candidate as training-time \textit{privileged context}~\cite{vapnik2009new, zhao2026self}, we obtain a candidate-conditioned query embedding, which we call a \textit{hindsight representation}.
This representation provides target-informed supervision for learning how to encode the query alone.

While the matched candidate provides useful context, learning an effective hindsight representation from it is non-trivial~\cite{yang2022toward,choudhury2025better}.
With direct access to the candidate, the hindsight representation may become overly influenced by candidate-specific information that is unavailable from the query alone~\cite{harne2026privileged}.
It may even collapse toward the candidate embedding, making the resulting supervision largely redundant with standard query--candidate matching.
The key challenge is therefore to leverage this privileged context while keeping the hindsight representation grounded in the query and transferable without candidate conditioning.

To this end, we introduce \textit{Hindsight-Enhanced Multimodal Embeddings} (\textbf{HEME}), a two-stage framework that first trains a hindsight teacher and then distills its supervision into a standalone embedder.
In the first stage, the hindsight teacher conditions query encoding on candidate context while limiting reliance on candidate-specific information.
Specifically, a hindsight attention mask allows candidate information to influence the embedding only through query content tokens, blocking direct candidate-to-readout access.
We also condition the hindsight pass on mismatched candidates, discouraging the teacher from simply following the provided candidate.
We further align intermediate query-token representations between the hindsight and query-only passes, encouraging candidate-conditioned representations to remain close to their query-only counterparts.
We then freeze the hindsight teacher and train a standalone student embedder by matching the teacher's hindsight embedding and similarity distribution.
At inference, the student encodes queries and candidates independently, without access to the matched candidate or the hindsight teacher.
Experimental results show that HEME achieves the highest overall MMEB-V2 scores among comparable models at both the 2B and 7B scales with single-pass inference, while remaining competitive on MMEB-V3 agent and text tasks without dedicated training data for these tasks.

Our main contributions can be summarized as follows:
\begin{itemize}
    \item We introduce hindsight supervision for universal multimodal embeddings, using matched candidates as training-time privileged context while retaining independent encoding at inference.
    \item We propose HEME, which learns query-grounded hindsight representations and distills their supervision into a standalone embedder.
    \item We show that HEME achieves superior overall performance on MMEB-V2 and competitive results on MMEB-V3 agent and text tasks, supported by extensive analyses of the proposed framework.
\end{itemize}

\section{Related Work}
\label{sec:related_work}

\textbf{Universal multimodal embeddings (UME).}
Embedding-based retrieval has progressed from contrastively trained dual encoders~\cite{karpukhin2020dense,radford2021learning} to instruction-aware retrievers capable of handling diverse retrieval tasks~\cite{wei2024uniir,zhang2024magiclens,lin2025mmembed,zhang2025bridging,liu2025lamra}.
With the rise of MLLM-based embedders, this direction has broadened into universal multimodal embeddings, where a single instruction-conditioned model supports heterogeneous inputs and a wide range of downstream tasks~\cite{jiang2024e5v,jiang2024vlm2vec,gu2025breaking,thirukovalluru2025breaking,meng2025vlm2vecv2,jian2025rzenembed,li2026qwen3vlembedding}.
More recently, reasoning-driven approaches enrich embedding construction through explicit reasoning traces~\cite{cui2025think,lan2026umer1,jiang2026embedrl,wu2026beyond}, latent reasoning~\cite{he2026plume,wu2026lame}, or adaptive reasoning~\cite{wang2026mmembr1,zhang2026thinkneeded}.
Unlike these inference-time reasoning approaches, HEME leverages the matched candidate only as training-time privileged context to refine query representations while preserving independent encoding at inference without additional reasoning steps.

\paragraphnew{Learning with hindsight.}
Learning with privileged information exploits signals available only during training~\cite{vapnik2009new,hoffman2016learning}, while generalized distillation extends this principle to teacher-student learning~\cite{lopez2016unifying}.
Recently, this paradigm has gained renewed attention in LLM self-distillation, where teachers conditioned on privileged context supervise students that must operate without it~\cite{zhao2026self,penaloza2026privileged,shenfeld2026self,hubotter2026reinforcement}.
For retrieval, prior work has similarly leveraged target-side information, often through teacher or posterior models, by incorporating reference outputs or images~\cite{paranjape2022hindsight,wen2023target}, relevant documents~\cite{xia2025momentum,chen2026efficient}, and future interactions~\cite{tang2026learning}.
Nonetheless, privileged conditioning can bias teacher supervision toward privileged content~\cite{harne2026privileged}, and its benefits depend on effective transfer to the student~\cite{yang2022toward,choudhury2025better}.
To mitigate such bias, HEME learns candidate-conditioned hindsight representations that remain grounded in the query and transfers their supervision to the final embedder.

\paragraphnew{Distillation for embedding models.}
Knowledge distillation~\cite{hinton2015distilling} is widely used in retrieval to transfer cross-encoder relevance scores to independently encoded retrievers~\cite{hofstatter2020improving,ren2021rocketqav2,miech2021thinking}, and has also been adopted in recent UME models~\cite{li2026umarvel,li2026qwen3vlembedding}.
Beyond score-level supervision, representation distillation aligns teacher and student embeddings or their relational structure~\cite{park2019relational,fang2021seed,kim2023embeddistill,yang2024clipkd,zhang2024jasper,vera2025embeddinggemma}.
In parallel, asymmetric retrieval methods distill between encoders of different capacities, typically supervising a lightweight query encoder with a stronger encoder used for offline indexing~\cite{budnik2021asymmetric,du2024image2sentence,kim2026towards,liu2026nanovdr}.
HEME instead distills across an information asymmetry rather than a capacity gap: teacher and student share the same architecture, but only the teacher receives the matched candidate as privileged context.

\begin{figure*}[t]
    \centering
    \includegraphics[width=0.97\textwidth]{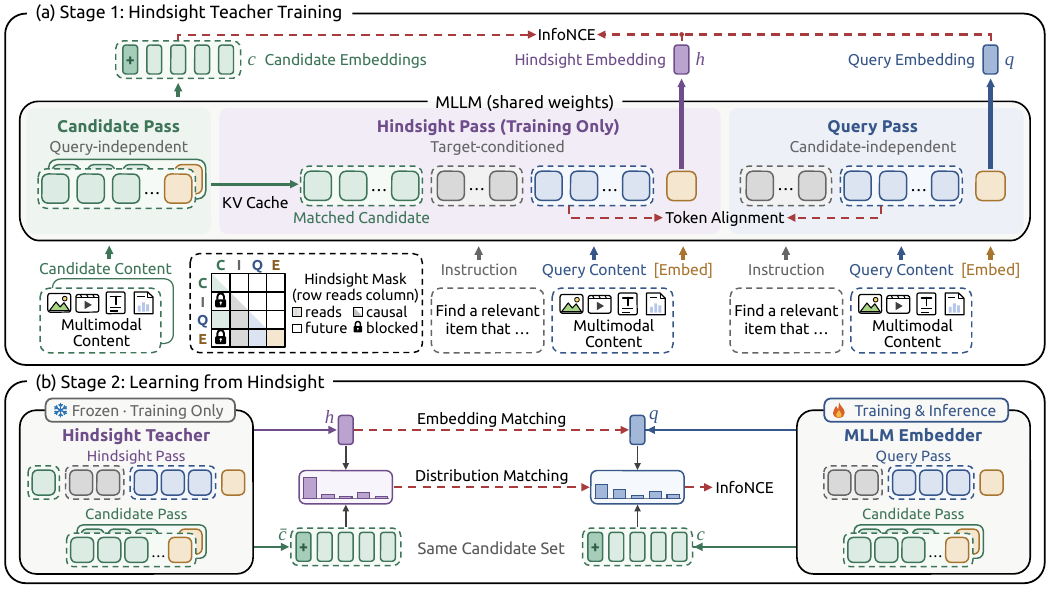}
    \vspace{-2mm}
    \caption{\textbf{Overview of HEME.}
    (a) \textbf{Hindsight teacher training.}
    The candidate and query passes encode their inputs independently, while the hindsight pass conditions the query on its matched candidate by reusing the candidate KV cache.
    The hindsight mask allows candidate context to influence the embedding only through query content tokens.
    Together with mismatched-candidate conditioning and query-token alignment, this encourages the hindsight representation to remain grounded in the query.
    The mismatched-candidate hindsight pass is omitted for clarity.
    (b) \textbf{Learning from hindsight.}
    The frozen hindsight teacher supervises a standalone student through embedding and similarity-distribution matching, while the student is also trained with standard InfoNCE.
    Distribution matching aligns the teacher and student similarity distributions over the same in-batch candidates, using their respective candidate embeddings ($\bar{\mathbf{c}}$ and $\mathbf{c}$).
    At inference, only the standalone student embedder is used.}
    \vspace{-3mm}
\label{fig:overview}
\end{figure*}

\section{Method}
\label{sec:method}

HEME learns a universal multimodal embedder that leverages privileged candidate context during training while preserving independent encoding at inference.
We first formulate MLLM-based embedding~(\cref{sec:prelim}), then introduce the hindsight representation and the challenges arising from candidate conditioning~(\cref{sec:hindsight}).
We then describe the two stages of HEME~(\cref{fig:overview}): training a hindsight teacher under constraints on candidate reliance~(\cref{sec:stage1}), and distilling its hindsight supervision into a standalone embedder that operates without candidate context~(\cref{sec:stage2}).

\subsection{Preliminaries}
\label{sec:prelim}

\paragraphnew{MLLM-based embedding.}
Each input query or candidate comprises multimodal content (\eg, text, images, videos, visual documents) paired with an optional task instruction.
Following recent universal embedders~\cite{jiang2024vlm2vec,meng2025vlm2vecv2}, we use a causal decoder-only MLLM and place a designated readout token \texttt{[Embed]} at the end of each input sequence.
We extract its last-layer hidden state and apply $\ell_2$-normalization to obtain the embedding vector.
Let $f_\theta$ denote the resulting embedder, and let $\mathbf{q}=f_\theta(Q)$ and $\mathbf{c}=f_\theta(C)$ denote the embeddings of a query sequence $Q$ and a candidate sequence $C$, respectively.
Because both embeddings are unit-normalized, their inner product $\mathbf{q}^{\top}\mathbf{c}$ equals their cosine similarity and serves as the relevance score.
Queries and candidates are encoded independently, allowing candidate embeddings to be precomputed and indexed offline.

\paragraphnew{Contrastive objective.}
Given a batch of $N$ matched query--candidate pairs $\{(Q_i, C_i)\}_{i=1}^{N}$, we adopt the standard InfoNCE loss~\cite{oord2018representation}.
For the query embedding $\mathbf{q}_i$, we define
\begin{equation}
    \mathcal{L}_{\text{NCE}} = \ell(\mathbf{q}_i) = -\log \frac{\exp(\mathbf{q}_i^{\top}\mathbf{c}_i/\tau)}{\sum_{j=1}^{N}\exp(\mathbf{q}_i^{\top}\mathbf{c}_j/\tau)},
    \label{eq:nce}
\end{equation}
where $\tau$ denotes the temperature.
For notational simplicity, all objectives are written for a single query $i$ and averaged over the batch.
We reuse $\ell(\cdot)$ below for other query-side representations by substituting them for $\mathbf{q}_i$ in \cref{eq:nce}.

\subsection{Hindsight Representation}
\label{sec:hindsight}

While the contrastive objective $\mathcal{L}_{\text{NCE}}$ supervises which candidate matches a query, it does not explicitly indicate which query cues underlie that match.
To provide such a signal during training, we construct a candidate-conditioned query representation by allowing the query to attend to its matched candidate.
Specifically, we prepend the candidate sequence $C_i$ to the query sequence $Q_i$ under causal attention and extract the representation at the query's terminal \texttt{[Embed]} token:
\begin{equation}
    \mathbf{h}_i = f_\theta([C_i; Q_i]),
    \label{eq:hindsight_repr}
\end{equation}
which we term the \emph{hindsight representation}.
Unlike the query-only embedding $\mathbf{q}_i$, $\mathbf{h}_i$ has access to the matched candidate as training-time privileged context~\cite{vapnik2009new}, while still serving as a query-side embedding scored against candidate embeddings.

Direct candidate conditioning, however, can make the hindsight representation overly dependent on candidate-specific information.
First, under standard causal attention, the terminal \texttt{[Embed]} token can directly attend to candidate tokens, allowing candidate information to dominate $\mathbf{h}_i$.
If $\mathbf{h}_i$ collapses toward the candidate embedding, it provides limited query-dependent supervision beyond the positive-pair signal already supplied by the contrastive objective.
Second, the hindsight representation may encode candidate-specific details that are unavailable from the query alone, making such supervision difficult to transfer to a query-only embedder.
We therefore seek a \emph{query-grounded} hindsight representation that benefits from candidate context while remaining transferable without candidate conditioning.

\subsection{Hindsight Teacher Training}
\label{sec:stage1}
To address these challenges, we train the hindsight teacher with three components: a hindsight mask and mismatched-candidate conditioning to prevent copying, and query-token alignment to keep representations grounded in the query.

\paragraphnew{Hindsight pass.}

The hindsight teacher is a single MLLM $f_\theta$ that performs candidate, query, and hindsight passes with shared weights.
The candidate and query passes independently compute $\mathbf{c}_i = f_\theta(C_i)$ and $\mathbf{q}_i = f_\theta(Q_i)$.
Meanwhile, the hindsight pass computes $\mathbf{h}_i$ as in \cref{eq:hindsight_repr} under the hindsight mask described below; hereafter, $\mathbf{h}_i$ denotes this masked representation.
Because causal attention makes the candidate-prefix computation independent of the subsequent query, we reuse the detached key--value cache from the candidate pass and process only the query sequence.
We train $\mathbf{h}_i$ with the same contrastive objective $\ell(\mathbf{h}_i)$ over the candidate embeddings $\{\mathbf{c}_j\}_{j=1}^{N}$.

\paragraphnew{Hindsight mask.}
To prevent direct candidate access to the embedding readout, the hindsight mask selectively restricts attention to the candidate prefix.
Only the query content tokens, denoted $\mathcal{T}_i$, can attend to candidate tokens; all other query-side tokens, including the task instruction and the readout \texttt{[Embed]} token, are blocked from the candidate prefix (Hindsight Mask in \cref{fig:overview}a).
This blocks both the direct candidate-to-\texttt{[Embed]} path and a potential shortcut through the instruction tokens.
As a result, candidate context can influence $\mathbf{h}_i$ only by modulating the representations of the query content tokens.

\paragraphnew{Mismatched candidates.}
Even under the hindsight mask, the query content tokens can still relay candidate information to \texttt{[Embed]}, allowing $\mathbf{h}_i$ to simply follow the provided candidate.
To discourage this behavior, we also condition the query on a mismatched candidate $C_k$ (an in-batch negative with $k \neq i$), yielding $\tilde{\mathbf{h}}_i = f_\theta([C_k; Q_i])$ under the same mask.
Despite being conditioned on this mismatched candidate $C_k$, $\tilde{\mathbf{h}}_i$ is trained to retrieve the matched candidate embedding $\mathbf{c}_i$:
\begin{equation}
    \mathcal{L}_{\text{H}} = \ell(\mathbf{h}_i) + \ell(\tilde{\mathbf{h}}_i).
    \label{eq:hindsight_loss}
\end{equation}
Since $C_k$ is a negative for $Q_i$, following the mismatched candidate would assign a high score to an incorrect candidate, which is directly penalized by $\ell(\tilde{\mathbf{h}}_i)$.

\paragraphnew{Token alignment.}
To further encourage the hindsight representation to remain grounded in the query, we align intermediate query-token representations between the hindsight and query passes, following prior work on intermediate representation alignment~\cite{yu2025representation,li2026spatial,yoon2025visual}.
Specifically, for the query content tokens $\mathcal{T}_i$, we minimize
\begin{equation}
    \mathcal{L}_{\text{align}} =
    \frac{1}{|\mathcal{T}_i|}
    \sum_{t\in\mathcal{T}_i}
    \Big[
    1-\cos\big(
    \mathbf{z}^{\text{H},m}_{i,t},\,
    \mathbf{z}^{\text{Q},m}_{i,t}
    \big)
    \Big],
    \label{eq:align}
\end{equation}
where $\mathbf{z}^{\text{H},m}_{i,t}$ and $\mathbf{z}^{\text{Q},m}_{i,t}$ are the hidden states of token $t$ at layer $m$ in the hindsight pass (with the matched candidate) and the query pass, respectively.
This alignment directly regularizes the differences between candidate-conditioned and query-only representations of the same query tokens.

\paragraphnew{Teacher objective.}
The hindsight teacher is trained with the standard contrastive objective, the hindsight objective, and the token-alignment objective:
\begin{equation}
    \mathcal{L}_{\text{teacher}}
    = \mathcal{L}_{\text{NCE}}
    + \mathcal{L}_{\text{H}}
    + \mathcal{L}_{\text{align}}.
    \label{eq:teacher}
\end{equation}
We stop gradients through reused candidate KV caches and embeddings, so the hindsight objectives backpropagate only through query-side computations.

\subsection{Learning from Hindsight}
\label{sec:stage2}

The hindsight teacher can use the matched candidate to produce a target-informed query representation, but this privileged context is unavailable at inference.
We therefore transfer the teacher's hindsight supervision to a standalone \emph{student} embedder that encodes queries and candidates independently~(\cref{fig:overview}b).
To provide fixed supervision during distillation, we freeze the trained teacher and train a separate student $f_\phi$, initialized from the same pretrained MLLM as the teacher.
Here, $\mathbf{q}_i = f_\phi(Q_i)$ and $\mathbf{c}_i = f_\phi(C_i)$ denote the student's embeddings, while the frozen teacher provides the hindsight representation $\mathbf{h}_i$, its query-only embedding $\bar{\mathbf{q}}_i = f_\theta(Q_i)$, and its candidate embedding $\bar{\mathbf{c}}_i = f_\theta(C_i)$.
The student is trained with three objectives: embedding matching, distribution matching, and the standard InfoNCE.

\paragraphnew{Embedding matching.}
We first transfer the hindsight representation by matching the student query embedding $\mathbf{q}_i$ to the frozen teacher representation $\mathbf{h}_i$:
\begin{equation}
    \mathcal{L}_{\text{emb}} = 1 - \cos(\mathbf{q}_i, \mathbf{h}_i) = 1 - \mathbf{q}_i^{\top}\mathbf{h}_i.
    \label{eq:emb_match}
\end{equation}
Since the frozen teacher was trained with constraints that encourage $\mathbf{h}_i$ to remain query-grounded, $\mathbf{h}_i$ provides a target-informed training signal for the student's query representation.

\paragraphnew{Distribution matching.}
While embedding matching aligns individual representations pointwise, it does not explicitly supervise the relative similarities between the query and different candidates.
We therefore match the similarity distributions induced by the teacher and student over the same in-batch candidate set.
Specifically, let $\mathbf{p}^{\text{T}}_i$ and $\mathbf{p}^{\text{S}}_i$ denote the probability distributions over all $N$ in-batch candidates, with $p^{\text{T}}_{i,j} \propto \exp(\mathbf{h}_i^{\top}\bar{\mathbf{c}}_j/\tau_{\text{d}})$ and $p^{\text{S}}_{i,j} \propto \exp(\mathbf{q}_i^{\top}\mathbf{c}_j/\tau_{\text{d}})$, where $\tau_{\text{d}}$ is the distillation temperature.
We minimize the Kullback--Leibler (KL) divergence between the teacher and student distributions:
\begin{equation}
    \mathcal{L}_{\text{dist}} = D_{\mathrm{KL}}\big(\mathbf{p}^{\text{T}}_i \,\|\, \mathbf{p}^{\text{S}}_i\big) = \sum_{j=1}^{N} p^{\text{T}}_{i,j} \log \frac{p^{\text{T}}_{i,j}}{p^{\text{S}}_{i,j}}.
    \label{eq:dist_loss}
\end{equation}
This objective transfers the hindsight teacher's similarity structure over the candidate set, providing relational supervision complementary to embedding matching.

\paragraphnew{Student objective.}
The student combines the contrastive loss $\mathcal{L}_{\text{NCE}} = \ell(\mathbf{q}_i)$, computed against its own candidate embeddings $\{\mathbf{c}_j\}_{j=1}^{N}$, with the two distillation objectives:
\begin{equation}
    \mathcal{L}_{\text{student}} = \alpha\mathcal{L}_{\text{NCE}} + \mathcal{L}_{\text{emb}} + \mathcal{L}_{\text{dist}},
    \label{eq:student_obj}
\end{equation}
where $\alpha$ controls the balance between the contrastive and distillation objectives, as in prior retrieval distillation~\cite{wang2022text,wang2023simlm}.
The contrastive objective preserves standard embedding training, while embedding and distribution matching provide pointwise and relational supervision, respectively.
Since the teacher and student share the same architecture and pretrained initialization, HEME distills across an \emph{information asymmetry} rather than a capacity gap: the teacher has privileged access to the matched candidate, whereas the student encodes the query alone.
At inference, only the standalone student embedder is used for independent query and candidate encoding.

\section{Experiments}
\label{sec:experiments}
\begin{table*}[t]
\centering
\caption{\textbf{Results on MMEB-V2~\cite{meng2025vlm2vecv2}.} 
HEME achieves the highest overall score in both the $\sim$2B and $\sim$7B comparison groups; models with larger backbones or external training data are reported separately as references.
We report Hit@1 for image and video tasks and NDCG@5 for visual document tasks.
Avg and All denote dataset averages within each category and overall, respectively. CLS: classification, QA: question answering, RET: retrieval, GD: grounding, MRET: moment retrieval, VDR: ViDoRe, VR: VisRAG, OOD: out-of-domain.}
\label{tab:main}
\vspace{-2mm}
\renewcommand{\arraystretch}{1.05}
\begin{adjustbox}{max width=\textwidth}
\begin{tabular}{l*{16}{>{\centering\arraybackslash}p{0.75cm}}}
\toprule
\multirow{2}{*}{\textbf{Model}} & \multicolumn{5}{c}{\textbf{Image}} & \multicolumn{5}{c}{\textbf{Video}} & \multicolumn{5}{c}{\textbf{VisDoc}} & \\
\cmidrule(lr){2-6}\cmidrule(lr){7-11}\cmidrule(lr){12-16}
 & CLS & QA & RET & GD & Avg & CLS & QA & RET & \makebox[\linewidth]{MRET} & Avg & \makebox[\linewidth]{VDRv1} & \makebox[\linewidth]{VDRv2} & VR & OOD & Avg & \textbf{All} \\
\midrule
\# Datasets & 10 & 10 & 12 & 4 & 36 & 5 & 5 & 5 & 3 & 18 & 10 & 4 & 6 & 4 & 24 & 78 \\
\midrule
\rowcolor{gray!10}
\multicolumn{17}{c}{\textit{Reference models (larger backbones / external training data)}} \\
RzenEmbed-v2-7B~\cite{jian2025rzenembed} & 70.6 & 71.7 & 78.5 & 92.1 & 75.9 & 58.8 & 63.5 & 51.0 & 45.5 & 55.7 & 89.7 & 60.7 & 88.7 & 44.4 & 77.1 & 71.6 \\
Qwen3-VL-Embed-8B~\cite{li2026qwen3vlembedding} & 74.2 & 81.1 & 80.2 & 92.3 & 80.1 & 78.4 & 71.0 & 58.7 & 56.1 & 67.1 & 87.2 & 69.9 & 88.7 & 73.3 & 82.4 & 77.8 \\
DME-9B~\cite{chen2026douyin} & 74.5 & 80.9 & 78.2 & 94.6 & 79.8 & 87.7 & 71.0 & 61.0 & 58.5 & 70.8 & 87.6 & 57.8 & 94.5 & 73.5 & 82.0 & 78.4 \\
WeMM-Embedding-9B~\cite{zhou2026wemm} & 76.2 & 82.2 & 81.7 & 95.6 & 81.9 & 87.4 & 77.7 & 67.4 & 58.5 & 74.3 & 90.5 & 65.1 & 90.9 & 71.8 & 83.3 & 80.6 \\
\midrule
\rowcolor{gray!10}
\multicolumn{17}{c}{\textit{$\sim$2B model size}} \\
GME~\cite{zhang2025bridging} & 54.4 & 29.9 & 66.9 & 55.5 & 51.9 & 34.9 & 42.0 & 25.6 & 32.4 & 33.9 & 86.1 & 54.0 & 82.5 & 43.1 & 72.7 & 54.1 \\
VLM2Vec-V2~\cite{meng2025vlm2vecv2} & 62.9 & 56.3 & 69.5 & 77.3 & 64.9 & 39.3 & 34.3 & 28.8 & 38.5 & 34.9 & 75.5 & 44.9 & 79.4 & 39.4 & 65.4 & 58.0 \\
UME-R1~\cite{lan2026umer1} & 64.8 & 62.8 & 67.6 & 77.2 & 66.6 & 44.3 & 51.2 & 32.9 & 39.7 & 42.2 & 72.4 & 46.2 & 79.2 & 37.2 & 63.9 & 60.1 \\
PLUME~\cite{he2026plume} & 66.5 & 59.2 & 67.6 & 79.7 & 66.3 & 45.0 & 52.3 & 33.5 & 46.7 & 44.1 & 72.1 & 49.8 & 78.1 & 57.4 & 67.5 & 61.6 \\
TTE$_s$~\cite{cui2025think} & 67.9 & 66.6 & 70.2 & 84.1 & 70.1 & 47.3 & 49.1 & 33.2 & 32.1 & 41.3 & 77.5 & 53.2 & 83.2 & 41.1 & 68.8 & 63.1 \\
RIME~\cite{wu2026beyond} & 67.9 & 64.4 & 69.8 & 82.1 & 69.1 & 48.0 & 52.1 & 33.6 & 39.2 & 43.7 & 76.4 & 51.4 & 81.7 & 63.9 & 71.4 & 64.1 \\
LaME~\cite{wu2026lame} & 67.6 & 66.2 & 70.5 & 81.2 & 69.3 & 45.6 & 52.3 & 36.3 & 43.5 & 44.5 & 78.2 & 48.4 & 82.8 & 55.4 & 72.1 & 64.4 \\
MMEmb-R1~\cite{wang2026mmembr1} & 64.5 & 68.1 & 70.0 & 88.9 & 70.0 & 56.3 & 52.8 & 47.6 & 42.5 & 50.6 & 74.1 & 56.0 & 78.9 & 48.3 & 68.0 & 65.0 \\
VaME~\cite{wu2026vame} & 67.6 & 66.6 & 70.0 & 80.3 & 69.5 & 46.2 & 54.8 & 38.2 & 45.5 & 46.3 & 79.4 & 55.2 & 85.6 & 64.1 & 74.4 & 65.7 \\
\rowcolor{Periwinkle!10}
\textbf{HEME (Ours)} & 66.3 & 66.5 & 71.4 & 81.6 & 69.7 & 49.4 & 53.2 & 37.1 & 39.2 & 45.3 & 84.9 & 55.3 & 87.3 & 65.7 & 77.3 & 66.4 \\
\midrule
\rowcolor{gray!10}
\multicolumn{17}{c}{\textit{$\sim$7B model size}} \\
GME~\cite{zhang2025bridging} & 57.7 & 34.7 & 71.2 & 59.3 & 56.0 & 37.4 & 50.4 & 28.4 & 38.2 & 38.6 & 89.4 & 55.6 & 85.0 & 44.4 & 75.2 & 57.8 \\
UME-R1~\cite{lan2026umer1} & 67.1 & 69.2 & 71.9 & 84.9 & 71.3 & 48.6 & 60.7 & 38.2 & 39.3 & 47.5 & 75.7 & 50.5 & 83.7 & 37.6 & 67.1 & 64.5 \\
TTE$_s$~\cite{cui2025think} & 69.7 & 72.4 & 74.0 & 90.6 & 74.2 & 49.1 & 60.6 & 36.4 & 37.2 & 46.8 & 84.1 & 62.7 & 91.9 & 47.6 & 76.4 & 68.6 \\
RIME~\cite{wu2026beyond} & 70.3 & 71.7 & 73.2 & 86.3 & 73.4 & 52.6 & 62.0 & 38.4 & 41.6 & 49.4 & 80.9 & 55.6 & 85.8 & 66.9 & 75.6 & 68.6 \\
LaME~\cite{wu2026lame} & 70.0 & 71.0 & 73.1 & 85.8 & 73.0 & 51.0 & 64.8 & 40.3 & 44.6 & 50.8 & 81.5 & 58.2 & 85.0 & 68.2 & 75.9 & 68.8 \\
MMEmb-R1~\cite{wang2026mmembr1} & 68.1 & 71.8 & 77.7 & 87.1 & 74.4 & 60.2 & 56.0 & 51.2 & 41.1 & 53.3 & 79.6 & 65.3 & 84.7 & 58.0 & 74.9 & 69.7 \\
VaME~\cite{wu2026vame} & 71.3 & 70.3 & 73.7 & 84.1 & 73.2 & 49.9 & 58.6 & 42.0 & 49.3 & 50.0 & 83.2 & 59.0 & 89.4 & 67.9 & 78.2 & 69.4 \\
\rowcolor{Periwinkle!10}
\textbf{HEME (Ours)} & 69.5 & 71.8 & 74.9 & 87.0 & 73.9 & 53.2 & 60.2 & 40.7 & 41.5 & 49.7 & 88.1 & 60.9 & 88.7 & 69.3 & 80.6 & 70.4 \\
\bottomrule
\end{tabular}
\end{adjustbox}
\vspace{-5mm}
\end{table*}

\subsection{Experimental Setup}
\paragraphnew{Benchmark and metrics.}
We evaluate HEME on MMEB-V2~\cite{meng2025vlm2vecv2}, a multimodal embedding benchmark comprising 78 datasets across image (36), video (18), and visual document (24) tasks.
Following the standard protocol, we report Hit@1 for image and video tasks and NDCG@5 for visual document tasks.
We further evaluate on the 47 agent and 53 text tasks of MMEB-V3~\cite{huang2026mmebv3}, reporting Hit@1 and NDCG@5, respectively.
We exclude the 11 audio tasks because Qwen2-VL does not support audio input.
Category scores are averaged across datasets within each category, and overall scores are averaged across all evaluated datasets within each benchmark.
Full benchmark details and task-level results are provided in Appendices~\ref{sec:supp_mmebv2} and~\ref{sec:supp_mmebv3}.

\paragraphnew{Implementation details.}
We build HEME on Qwen2-VL-2B and Qwen2-VL-7B~\cite{wang2024qwen2vl}.
Our training mixture comprises image-based vision-language tasks from MMEB-train~\cite{jiang2024vlm2vec}, video-language instructions from LLaVA-Hound~\cite{zhang2024direct}, and visual document retrieval from ViDoRe~\cite{faysse2024colpali} and VisRAG~\cite{yu2024visrag}.
Initialized from the same pretrained MLLM, the teacher and student are each trained for one epoch with a global batch size of 512 on NVIDIA H200 GPUs.
We use AdamW with a learning rate of $1\times10^{-5}$ ($5\times10^{-6}$ for 7B), a weight decay of 0.01, and 100 linear warmup steps followed by cosine decay.
We freeze the vision encoder and fully fine-tune the multimodal projector and language decoder.
We set the contrastive temperature to $\tau=0.02$, the distillation temperature to $\tau_{\text{d}}=0.07$, and the balance weight to $\alpha=0.2$.
We apply query-token alignment at layer $m=21$ of the 28-layer language decoder.
For each query, we uniformly sample one mismatched candidate from the in-batch negatives.
All ablation studies use the 2B backbone and are conducted on MMEB-V2.

\begin{table}[t]
\centering
\caption{\textbf{Inference efficiency.}
Latency is measured on MMEB-V2 samples using a single NVIDIA H100 with batch size 1.
Steps denotes the average number of sequential reasoning steps; $^\dagger$LaME processes eight latent tokens within a single forward pass.}
\label{tab:cost}
\vspace{-2mm}
\footnotesize
\renewcommand{\arraystretch}{1.15}
\setlength{\tabcolsep}{5.7pt}
\begin{tabular}{lcrrr}
\toprule
Method & Reasoning & Steps & Latency (ms) & MMEB-V2 \\
\midrule
UME-R1~\cite{lan2026umer1} & Explicit & 408 & 10{,}903 & 60.1 \\
RIME~\cite{wu2026beyond}   & Explicit & 153 & 4{,}145  & 64.1 \\
PLUME~\cite{he2026plume}   & Latent   & 8   & 295      & 61.6 \\
LaME~\cite{wu2026lame}     & Latent   & 0\rlap{$^\dagger$} & 51 & 64.4 \\
\midrule
\rowcolor{Periwinkle!10}
\textbf{HEME (Ours)}       & -- & 0 & 49 & {66.4} \\
\bottomrule
\end{tabular}
\vspace{-5mm}
\end{table}

\subsection{Main Results}
\label{sec:main_results}

\paragraphnew{Baselines.}
For MMEB-V2, we compare HEME with models built on the same Qwen2-VL backbones at the 2B and 7B scales.
These include standard UME baselines (GME~\cite{zhang2025bridging}, VLM2Vec-V2~\cite{meng2025vlm2vecv2}) and reasoning-driven models using explicit generation (UME-R1~\cite{lan2026umer1}, TTE$_s$~\cite{cui2025think}, RIME~\cite{wu2026beyond}), latent reasoning (PLUME~\cite{he2026plume}, LaME~\cite{wu2026lame}, VaME~\cite{wu2026vame}), and adaptive reasoning (MMEmb-R1~\cite{wang2026mmembr1}).
We additionally report models with larger backbones or external training data~\cite{jian2025rzenembed,li2026qwen3vlembedding,chen2026douyin,zhou2026wemm} as separate references.
MMEB-V2 baseline scores are taken from the original papers and the benchmark report~\cite{meng2025vlm2vecv2}.
For MMEB-V3, we use the baseline scores reported by the benchmark~\cite{huang2026mmebv3}.

\paragraphnew{MMEB-V2.}
Table~\ref{tab:main} reports the results on MMEB-V2.
HEME achieves the highest overall score in both the $\sim$2B and $\sim$7B groups, reaching 66.4 and 70.4, respectively, and outperforming the strongest baseline in each group by 0.7 points.
HEME also achieves the best visual-document performance in both groups, with averages of 77.3 and 80.6.
These results are achieved while retaining single-pass inference without additional test-time reasoning.

\begin{table}[t]
\centering
\caption{\textbf{Results on the agent and text tasks of MMEB-V3~\cite{huang2026mmebv3}.}
HEME achieves the highest agent score at the 7B scale and the best overall scores among Qwen2-VL-based models.
We report Hit@1 for agent and NDCG@5 for text tasks; All denotes the average over all 100 datasets.
$^\dagger$Models use additional text-only training data.
Per-subtask results are in Appendix~\ref{sec:supp_mmebv3}.}
\label{tab:mmebv3}
\vspace{-2mm}
\setlength{\tabcolsep}{5pt}
\resizebox{\linewidth}{!}{
\begin{tabular}{llccc}
\toprule
\textbf{Model} & \textbf{Backbone} & \textbf{Agent} & \textbf{Text} & \textbf{All} \\
\midrule
\# Datasets & & 47 & 53 & 100 \\
\midrule
VLM2Vec-V2~\cite{meng2025vlm2vecv2}
& Qwen2-VL-2B & 28.7 & 24.5 & 26.5 \\
E5-Omni$^{\dagger}$~\cite{chen2026e5omni}
& Qwen2.5-Omni-3B & 36.9 & 26.7 & 31.5 \\
Omni-Embed-Nemotron$^{\dagger}$~\cite{xu2025omniembednemotron}
& Qwen2.5-Omni-3B & 36.5 & 39.2 & 37.9 \\
Qwen3-VL-Embedding$^{\dagger}$~\cite{li2026qwen3vlembedding}
& Qwen3-VL-2B & 39.3 & 39.2 & 39.2 \\
\rowcolor{Periwinkle!10}
\textbf{HEME (Ours)}
& Qwen2-VL-2B & 36.9 & 34.4 & 35.6 \\
\midrule
VLM2Vec~\cite{jiang2024vlm2vec}
& Qwen2-VL-7B & 19.7 & 22.2 & 21.0 \\
LCO-Embedding-Omni$^{\dagger}$~\cite{xiao2025scaling}
& Qwen2.5-Omni-7B & 27.8 & 32.4 & 30.2 \\
E5-Omni$^{\dagger}$~\cite{chen2026e5omni}
& Qwen2.5-Omni-7B & 36.7 & 26.9 & 31.5 \\
GME$^{\dagger}$~\cite{zhang2025bridging}
& Qwen2-VL-7B & 35.6 & 37.1 & 36.4 \\
Qwen3-VL-Embedding$^{\dagger}$~\cite{li2026qwen3vlembedding}
& Qwen3-VL-8B & 38.4 & 42.5 & 40.6 \\
\rowcolor{Periwinkle!10}
\textbf{HEME-7B (Ours)} & Qwen2-VL-7B & 39.8 & 37.7 & 38.7 \\
\bottomrule
\end{tabular}}
\vspace{-6mm}
\end{table}

\paragraphnew{Inference efficiency.}
\Cref{tab:cost} compares the inference latency of HEME with reasoning-driven baselines among $\sim$2B models.
We measure the mean per-input latency on a single NVIDIA H100 with batch size 1, using 10 matched query--candidate pairs sampled from each MMEB-V2 dataset.
Unlike methods that require explicit generation or iterative latent reasoning, HEME produces the final embedding in a single forward pass without additional reasoning steps.
HEME requires 49\,ms, comparable to the single-pass latent baseline LaME (51\,ms), and is about 6$\times$, 85$\times$, and 223$\times$ faster than PLUME, RIME, and UME-R1, respectively.
At essentially the same latency, HEME achieves a higher MMEB-V2 score than LaME (66.4 vs. 64.4).
Efficient encoding is particularly important for large-scale embedding systems, where millions of items may need to be processed.
This consideration is especially relevant to MMEB-V3, whose million-scale corpora can make reasoning-based encoding computationally demanding~\cite{cui2026moemb}.

\paragraphnew{MMEB-V3.}
We further evaluate HEME on the agent and text tasks of MMEB-V3 (\cref{tab:mmebv3}).
HEME-7B achieves the highest agent score in the table (39.8).
At both the 2B and 7B scales, HEME also obtains the best overall scores among the Qwen2-VL-based baselines, reaching 35.6 and 38.7, respectively.
Scaling from 2B to 7B improves both agent and text performance.
Notably, these results are achieved without dedicated agent-specific or text-only training data, while remaining competitive overall with models trained on additional text-to-text retrieval data.

\begingroup
\begin{table}[t]
\centering
\caption{\textbf{Ablation of hindsight teacher design.}
Effects of the Stage~1 constraints on teacher diagnostics and student performance.
Copy$_{\text{rand}}$ / Copy$_{\text{hard}}$ (\%): rate of ranking the provided mismatched candidate first.
Mismatched candidates are either random in-batch negatives or hard negatives mined with the InfoNCE baseline.
$\cos(\mathbf{h},\bar{\mathbf{q}})$: similarity between hindsight and query-only embeddings.
Student All: overall MMEB-V2 score after Stage~2.}
\label{tab:abl_teacher}
\vspace{-2mm}
\footnotesize
\renewcommand{\arraystretch}{1.1}
\setlength{\tabcolsep}{3.0pt}
\begin{tabular}{@{}lcccc@{}}
\toprule
\multirow{2}{*}{\textbf{Teacher design}} & \multicolumn{3}{c}{\textbf{Teacher}} & \textbf{Student} \\
\cmidrule(lr){2-4}\cmidrule(lr){5-5}
 & Copy$_{\text{rand}}$\,($\downarrow$) & Copy$_{\text{hard}}$\,($\downarrow$) & $\cos(\mathbf{h},\bar{\mathbf{q}})$ & All \\
\midrule
InfoNCE (no teacher) & \cellcolor{Gray!20}-- & \cellcolor{Gray!20}-- & \cellcolor{Gray!20}-- & 63.7 \\
\midrule
\multicolumn{1}{@{}>{\columncolor{Periwinkle!10}[0pt][\tabcolsep]}l}{\textbf{HEME}} & \cellcolor{Periwinkle!10}0.4 & \cellcolor{Periwinkle!10}6.3 & \cellcolor{Periwinkle!10}0.93 & \multicolumn{1}{>{\columncolor{Periwinkle!10}[\tabcolsep][0pt]}c@{}}{66.4} \\
\, \textit{w/o} hindsight mask & 3.2 & 13.4 & 0.89 & 65.8 \\
\, \textit{w/o} token alignment & 2.8 & 9.2 & 0.79 & 65.4 \\
\, \textit{w/o} mismatched cand. & 67.6 & 89.5 & 0.72 & 63.1 \\
\, \textit{w/o} all constraints & 97.6 & 99.2 & 0.59 & 62.7 \\
\bottomrule
\end{tabular}
\vspace{-5mm}
\end{table}
\endgroup

\subsection{Ablation Studies}
\label{sec:ablations}

\paragraphnew{Hindsight teacher design.}
We ablate the constraints used to train the hindsight teacher and evaluate both teacher behavior and distilled student performance (\cref{tab:abl_teacher}).
We measure how often the teacher ranks a provided mismatched candidate first (copy rate), the similarity between hindsight and query-only embeddings, and the score of the student distilled from each teacher.
We compute teacher diagnostics during student training in the same in-batch setting used for hindsight supervision.
To limit evaluation overhead, we use 200 training steps ($\sim$102k queries), with hard mismatches drawn from the baseline's top-8 in-batch negatives.
All teacher variants are evaluated on the same batches; copy rates are computed over the in-batch candidate pool, while hindsight--query similarity is computed using the matched-candidate hindsight representation.

Removing the hindsight mask increases the hard-negative copy rate from 6.3\% to 13.4\%, whereas removing token alignment lowers hindsight--query similarity from 0.93 to 0.79.
The corresponding student scores decrease from 66.4 to 65.8 and 65.4, respectively.
Without mismatched-candidate conditioning, copy rates rise to 67.6\% and 89.5\% even with the mask and token alignment in place, and the student score drops to 63.1, below the InfoNCE baseline of 63.7.
Thus, the mask and token alignment alone do not prevent copying, as candidate information can still reach $\mathbf{h}$ through the query content tokens; mismatched-candidate conditioning plays a key role in suppressing this behavior.
Removing all constraints further increases both copy rates to over 97\% and reduces the student score to 62.7.
These results suggest that hindsight supervision is effective when the teacher draws on candidate context without letting it dominate the query representation.

\begin{table}[t]
\centering
\caption{\textbf{Ablation of hindsight transfer.}
Effects of the Stage~2 objectives on downstream performance.
All variants use the same hindsight teacher and retain the standard InfoNCE objective.}
\label{tab:abl_transfer}
\vspace{-2mm}
\footnotesize
\renewcommand{\arraystretch}{1.15}
\setlength{\tabcolsep}{7.7pt}
\begin{tabular}{lcccc}
\toprule
\textbf{Method} & Image & Video & VisDoc & All \\
\midrule
InfoNCE & 68.2 & 42.4 & 73.0 & 63.7 \\
\midrule
\,$+\,\mathcal{L}_{\text{emb}}$ & 69.4 & 43.0 & 76.8 & 65.6 \\
\,$+\,\mathcal{L}_{\text{dist}}$ & 69.5 & 44.3 & 76.9 & 65.9 \\
\rowcolor{Periwinkle!10}\,$+\,\mathcal{L}_{\text{emb}}+\mathcal{L}_{\text{dist}}$ (\textbf{HEME}) & 69.7 & 45.3 & 77.3 & 66.4 \\
\bottomrule
\end{tabular}
\vspace{-2mm}
\end{table}

\paragraphnew{Hindsight transfer.}
\Cref{tab:abl_transfer} examines how the teacher's hindsight supervision is transferred to the student.
Embedding matching and distribution matching individually improve the InfoNCE baseline from 63.7 to 65.6 and 65.9, respectively.
Their scores are close on image and visual-document tasks, while $\mathcal{L}_{\text{dist}}$ yields a higher video score (44.3 vs.\ 43.0).
Combining the two objectives raises the overall score to 66.4, outperforming either objective alone across image, video, and visual-document tasks, suggesting that the two forms of supervision are not fully redundant.

\begingroup
\begin{table}[t]
\centering
\caption{\textbf{Ablation of distillation target.}
Effects of the teacher and target representation used for Stage~2.
$\mathbf{q}$: query embedding of the InfoNCE-only teacher;
$\bar{\mathbf{q}}$: query-only embedding of the hindsight teacher;
$\mathbf{h}$: hindsight embedding.
$\times$: direct evaluation.}
\label{tab:abl_target}
\vspace{-2mm}
\footnotesize
\renewcommand{\arraystretch}{1.15}
\setlength{\tabcolsep}{6.9pt}
\begin{tabular}{l@{\hspace{\tabcolsep}\hspace{3pt}}c@{\hspace{3pt}\hspace{\tabcolsep}}cccc}
\toprule
\textbf{Teacher} & \textbf{Target} & Image & Video & VisDoc & All \\
\midrule
\multirow{2}{*}{{InfoNCE only}} & $\times$ & 68.2 & 42.4 & 73.0 & 63.7 \\
& $\mathbf{q}$ & 67.6 & 42.0 & 72.3 & 63.1 \\
\midrule
\multirow{3}{*}{{Hindsight (Stage~1)}}
& $\times$ & 68.7 & 43.2 & 75.8 & 65.0 \\
& $\bar{\mathbf{q}}$ & 68.8 & 43.4 & 76.4 & 65.3 \\
& \cellcolor{Periwinkle!10}$\mathbf{h}$
& \cellcolor{Periwinkle!10}69.7
& \cellcolor{Periwinkle!10}45.3
& \cellcolor{Periwinkle!10}77.3
& \cellcolor{Periwinkle!10}66.4 \\
\bottomrule
\end{tabular}
\vspace{-4mm}
\end{table}
\endgroup

\begin{figure*}[t]
    \centering
    \includegraphics[width=0.975\textwidth]{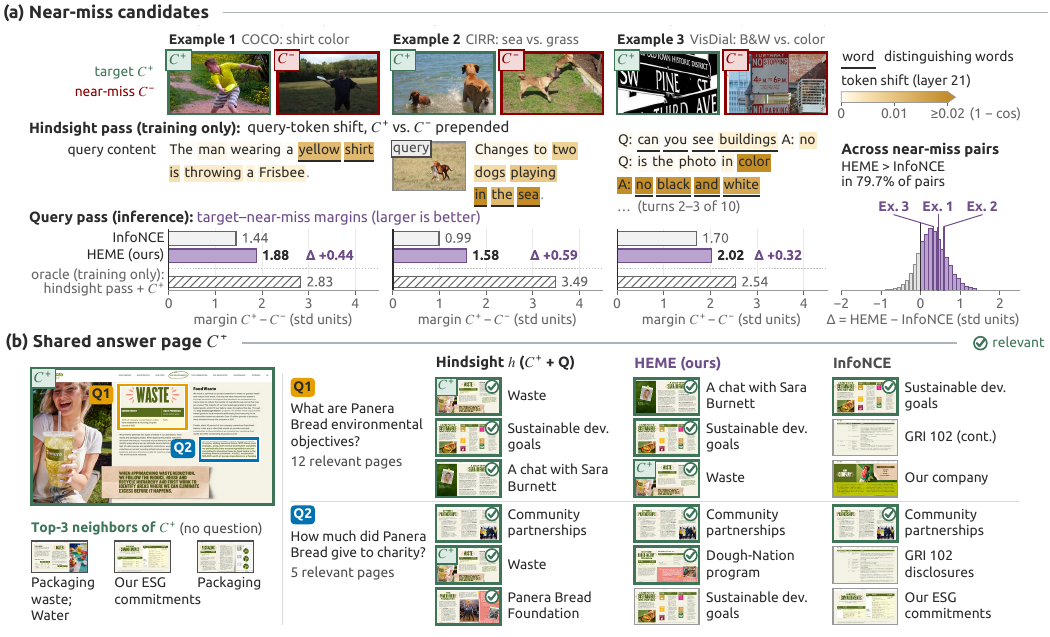}
    \vspace{-2.5mm}
    \caption{\textbf{Query-grounded hindsight and its transfer to inference.}
    (a)~For a target $C^{+}$ and a challenging near-miss $C^{-}$, the hindsight pass shows larger query-token shifts on words that distinguish the two candidates; at inference, HEME exhibits stronger target--near-miss separation in the illustrated examples.
    (b)~For a single page $C^{+}$ answering two different questions, conditioning on the same answer produces question-specific hindsight rankings; from the question alone, HEME exhibits corresponding retrieval behavior.}
    \label{fig:qualitative}
    \vspace{-4mm}
\end{figure*}

\paragraphnew{Distillation target.}
\Cref{tab:abl_target} examines whether the student gain comes from distillation itself, from a stronger teacher, or specifically from the hindsight target.
Distilling the query embedding of an InfoNCE-only teacher does not improve over the InfoNCE baseline, yielding 63.1 compared with 63.7.
The query pass of the hindsight teacher is already stronger than the baseline (65.0 vs.\ 63.7), and distilling its query-only representation $\bar{\mathbf{q}}$ yields 65.3.
Holding the same teacher fixed, distilling the hindsight representation $\mathbf{h}$ instead yields 66.4, a further 1.1-point improvement over $\bar{\mathbf{q}}$.
This suggests that the candidate-conditioned hindsight target provides additional supervision beyond the teacher's query-only representation.

\paragraphnew{Hindsight behavior.}
\Cref{fig:qualitative} examines how candidate context changes hindsight representations and whether corresponding behavior appears in the query-only student.
In \cref{fig:qualitative}(a), we hold the query fixed and compare the query-token hidden states when conditioned on the target or on a near-miss candidate selected from the top-10 hard negatives retrieved by the teacher's query pass.
In the illustrated examples, the largest shifts occur on words that distinguish the matched candidate from the near-miss candidate.
These distinctions span different levels of visual semantics, including a local object attribute (shirt color), scene context (sea vs.\ grass), and global image appearance (color vs.\ black-and-white).
We next evaluate whether these distinctions carry over to the student by measuring the query's separation between the matched candidate and the same near-miss.
At inference, HEME yields larger standardized target--near-miss margins than the InfoNCE baseline in all three examples, with the margin normalized by the standard deviation of each model's candidate similarities; the target-conditioned hindsight pass yields even larger margins (hatched bars).
Beyond these examples, HEME achieves a larger standardized margin than the baseline in 79.7\% of over 18k near-miss pairs from COCO, VisualNews, CIRR, and VisDial.

\Cref{fig:qualitative}(b) provides a complementary analysis of VisRAG page retrieval by holding the matched page fixed and varying the question.
The hindsight rankings differ from the nearest neighbors of the page alone, indicating that they are not determined solely by the candidate's embedding neighborhood.
Moreover, conditioning on the same page produces different rankings for the two questions: pages related to sustainability are favored for the environmental question, whereas pages about community programs and charitable activities are favored for the charity question.
From the question alone, HEME shows query-specific retrieval preferences similar to those of the hindsight rankings.
These analyses suggest that the hindsight teacher uses candidate context in a query-dependent manner and that the standalone student exhibits related behavior.

\section{Conclusion}
\label{sec:conclusion}
We presented HEME, a framework that uses matched candidates as training-time privileged context to learn universal multimodal embeddings while retaining independent encoding at inference.
HEME learns query-grounded hindsight representations by controlling reliance on the privileged context and distills them into a standalone embedder.
Experiments on MMEB-V2 and MMEB-V3 demonstrate the effectiveness of HEME across diverse tasks, with efficient inference and no additional reasoning steps.
Our analyses show that controlling this reliance is critical and that hindsight targets provide additional supervision beyond query-only teacher representations.

\section*{LLM Usage Statement}
We used ChatGPT and Codex (OpenAI) and Claude (Anthropic) to assist with manuscript editing, code implementation and debugging, and experimental design.
All AI-assisted outputs were reviewed and verified by the authors, who take full responsibility for the scientific claims, citations, code, experimental results, and final manuscript.

\section*{Acknowledgements}
This work was supported in part by the “Advanced GPU Utilization Support Program” funded by the Government of the Republic of Korea (Ministry of Science and ICT).
{
    \small
    \bibliographystyle{ieeenat_fullname}
    \bibliography{main}
}

\onecolumn
\appendix
\hypersetup{bookmarksnumbered=true}
\renewcommand{\thefigure}{S\arabic{figure}}
\renewcommand{\thetable}{S\arabic{table}}
\raggedbottom
\newcolumntype{Y}{>{\centering\arraybackslash}X}
\captionsetup[table]{position=top,skip=4pt}
\captionsetup[figure]{skip=5pt}
\setlength{\intextsep}{8pt plus 2pt minus 2pt}
\setcounter{figure}{0}
\setcounter{table}{0}

\section{Additional Qualitative Analysis}
\vspace{-2mm}
We provide additional qualitative examples on image (\cref{fig:supp_image}), visual document (\cref{fig:supp_visdoc}), and video (\cref{fig:supp_video}) retrieval.

\begingroup
\setlength{\intextsep}{2pt}
\captionsetup[figure]{skip=3pt}

\begin{figure}[H]
\centering
\includegraphics[width=0.9\linewidth]{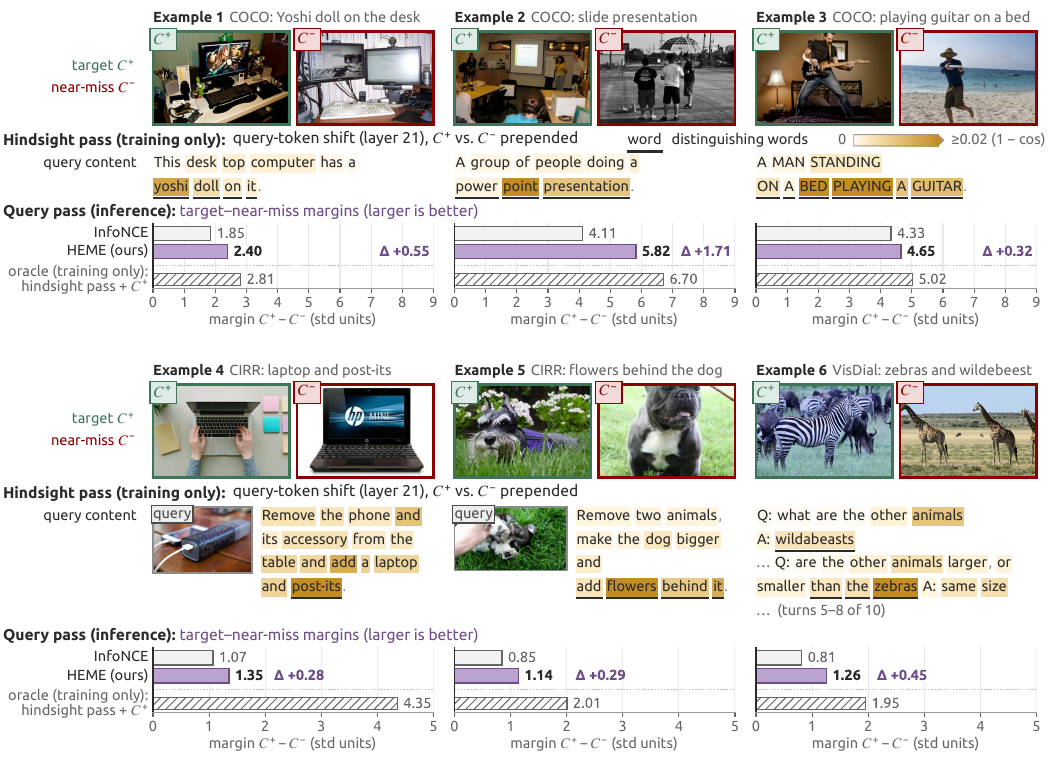}
\vspace{-1mm}
\caption{\textbf{Qualitative analysis of hindsight on image tasks.} Shading shows the query-token shift between conditioning on $C^+$ and $C^-$. Bars show retrieval margins normalized by the standard deviation of each model's candidate scores.}
\label{fig:supp_image}
\end{figure}

\begin{figure}[H]
\centering
\includegraphics[width=0.9\linewidth]{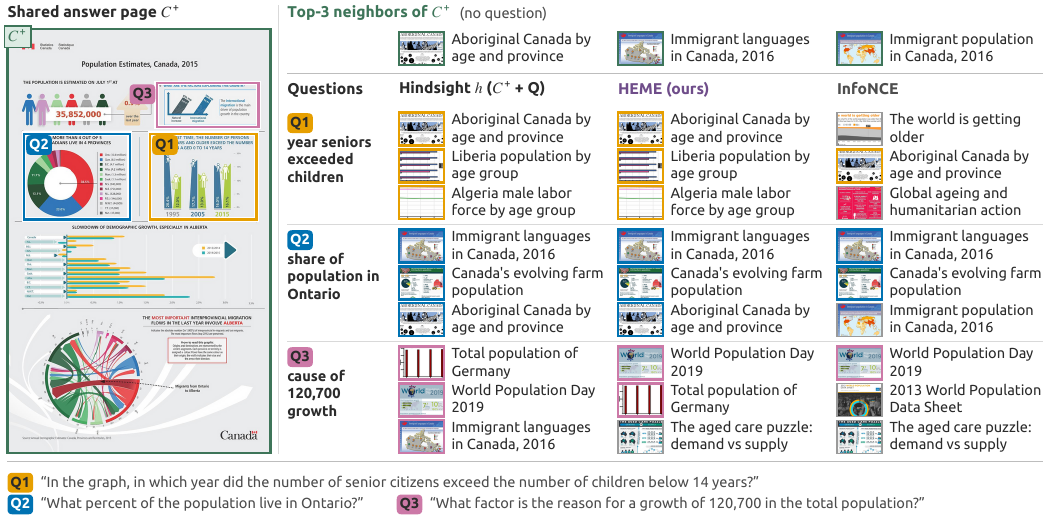}
\vspace{-1mm}
\caption{\textbf{Question-conditioned hindsight on visual document retrieval.} Three questions share the matched page $C^+$. For each, we show the top three pages from the hindsight pass, HEME, and InfoNCE, with $C^+$ omitted for clarity; HEME and InfoNCE use the question alone. Colored borders mark pages that also appear in the hindsight top three.}
\label{fig:supp_visdoc}
\end{figure}

\clearpage
\paragraphnew{Video retrieval.}
\Cref{fig:supp_video} extends the same token-shift analysis to text-to-video retrieval on MSVD, VATEX, YouCook2, and DiDeMo.

\begin{figure}[H]
\centering
\includegraphics[width=0.95\linewidth]{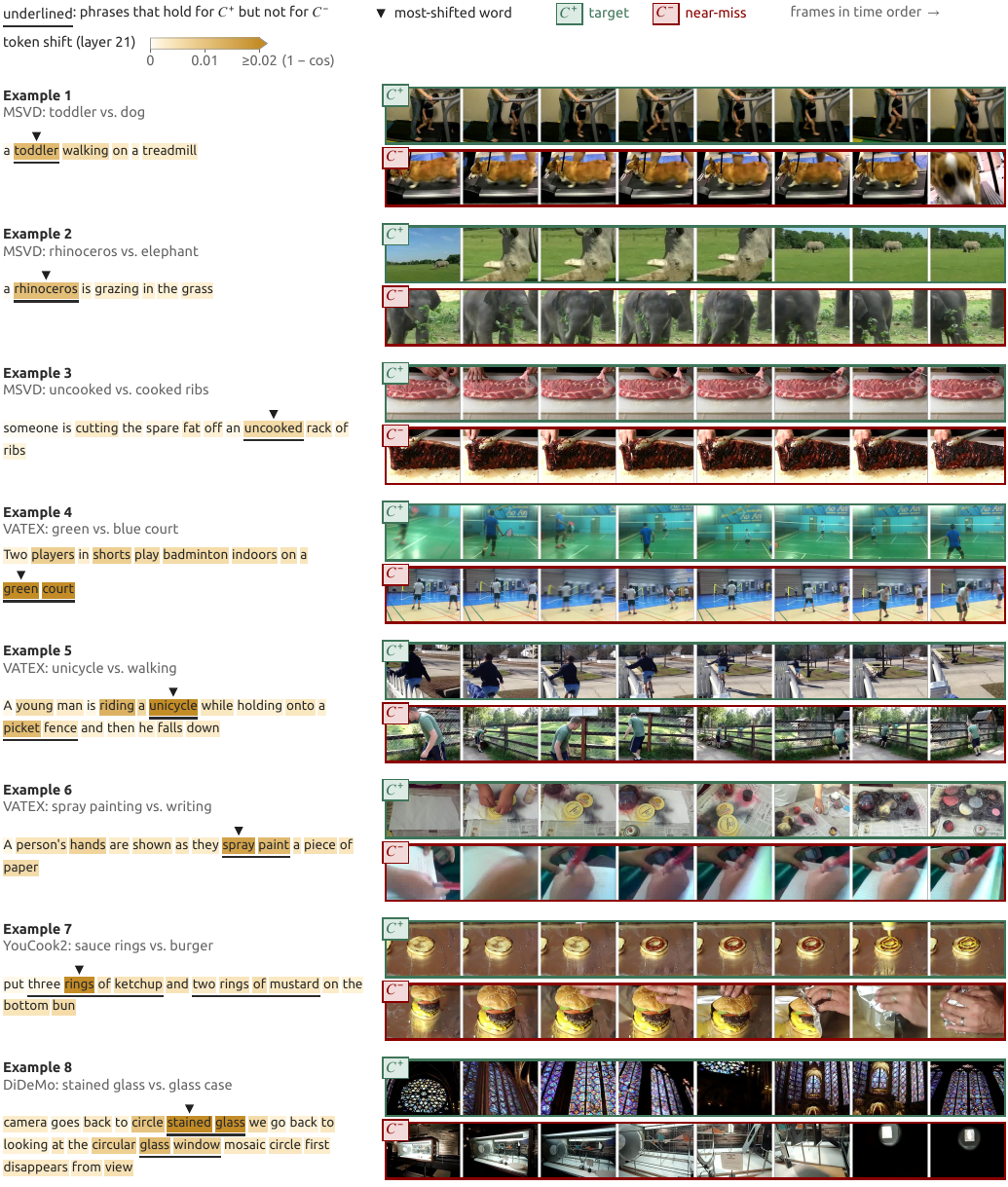}
\caption{\textbf{Qualitative analysis of hindsight on text-to-video retrieval.} Shading shows the query-token shift between conditioning on $C^+$ and $C^-$. Each pair shows frames from $C^+$ and $C^-$ in temporal order. Underlining marks target-specific phrases, and the triangle marks the word with the largest token shift.}
\label{fig:supp_video}
\end{figure}
\endgroup

\clearpage

\section{MMEB-V2 Evaluation}
\label{sec:supp_mmebv2}
\paragraphnew{Benchmark and metrics.}
We follow the MMEB-V2 evaluation protocol~\cite{meng2025vlm2vecv2}. We encode queries and candidates independently and rank candidates by their similarity to the query. Each dataset has equal weight in its category average and in the overall score.
\Cref{tab:supp_mmebv2} lists the 78 datasets by modality and task.
We report Hit@1 for the 36 image and 18 video datasets, and NDCG@5 for the 24 visual document (VisDoc) datasets.
For classification, question answering, and visual grounding, the candidates are class labels, answers, and image regions, respectively.

\begin{table}[H]
\centering
\caption{\textbf{MMEB-V2 datasets by modality and task.} The abbreviations t2i and i2t denote text-to-image and image-to-text retrieval.}
\label{tab:supp_mmebv2}
\small
\renewcommand{\arraystretch}{1.09}
\setlength{\tabcolsep}{4pt}
\begin{tabularx}{\linewidth}{@{}llc>{\raggedright\arraybackslash}X@{}}
\toprule
\textbf{Modality} & \textbf{Meta-task} & \textbf{\#} & \textbf{Datasets} \\
\midrule
Image & Classification (CLS) & 10 & ImageNet-1K, N24News, HatefulMemes, VOC2007, SUN397, Place365, ImageNet-A, ImageNet-R, ObjectNet, Country211 \\
 & Question answering (QA) & 10 & OK-VQA, A-OKVQA, DocVQA, InfographicsVQA, ChartQA, Visual7W, ScienceQA, VizWiz, GQA, TextVQA \\
 & Retrieval (RET) & 12 & VisDial, CIRR, VisualNews (t2i), VisualNews (i2t), MSCOCO (t2i), MSCOCO (i2t), NIGHTS, WebQA, FashionIQ, Wiki-SS-NQ, OVEN, EDIS \\
 & Grounding (GD) & 4 & MSCOCO, RefCOCO, RefCOCO-Matching, Visual7W-Pointing \\
\midrule
Video & Classification (CLS) & 5 & K700, SSv2, HMDB51, UCF101, Breakfast \\
 & Question answering (QA) & 5 & MVBench, Video-MME, NExT-QA, EgoSchema, ActivityNetQA \\
 & Retrieval (RET) & 5 & MSR-VTT, MSVD, DiDeMo, VATEX, YouCook2 \\
 & Moment retrieval (MRET) & 3 & QVHighlights, Charades-STA, MomentSeeker \\
\midrule
VisDoc & ViDoRe v1 (VDRv1) & 10 & ArxivQA, DocVQA, InfoVQA, TabFQuAD, TAT-DQA, ShiftProject, SyntheticDocQA (AI, Energy, Government Reports, Healthcare Industry) \\
 & ViDoRe v2 (VDRv2) & 4 & ESG Reports (human-labeled), plus multilingual Biomedical Lectures, Economics Reports, and ESG Reports \\
 & VisRAG (VR) & 6 & ArxivQA, ChartQA, MP-DocVQA, SlideVQA, InfoVQA, PlotQA \\
 & Out-of-domain (OOD) & 4 & ViDoSeek (page, doc), MMLongBench-Doc (page, doc) \\
\midrule
\multicolumn{2}{@{}l}{Total} & 78 & \\
\bottomrule
\end{tabularx}
\end{table}

\paragraphnew{Hyperparameter analysis.}
For Qwen2-VL-2B, we vary one parameter at a time (\cref{tab:supp_hyperparameter}).
We keep the teacher fixed when varying the contrastive loss weight $\alpha$ or the distillation temperature $\tau_d$.
When varying the token alignment layer $m$, we retrain both teacher and student. The default setting ($\alpha=0.2$, $\tau_d=0.07$, $m=21$) achieves the highest overall score in each sweep.

\begin{table}[H]
\centering
\caption{\textbf{Hyperparameter analysis on MMEB-V2.} Shading marks the default settings. Bold marks the highest score in each column of each sweep.}
\label{tab:supp_hyperparameter}
\footnotesize
\renewcommand{\arraystretch}{1.1}
\setlength{\tabcolsep}{3pt}
\begin{subtable}[t]{0.33\textwidth}
\centering
\caption{Balance weight $\alpha$}
\begin{tabular}{lcccc}
\toprule
$\alpha$ & Image & Video & VisDoc & All \\
\midrule
0.1 & 69.4 & \textbf{46.0} & 76.4 & 66.1 \\
\rowcolor{Periwinkle!10}
0.2 & \textbf{69.7} & 45.3 & \textbf{77.3} & \textbf{66.4} \\
1.0 & 68.7 & 43.5 & 76.7 & 65.3 \\
\bottomrule
\end{tabular}
\end{subtable}
\hfill
\begin{subtable}[t]{0.33\textwidth}
\centering
\caption{Distillation temperature $\tau_d$}
\begin{tabular}{lcccc}
\toprule
$\tau_d$ & Image & Video & VisDoc & All \\
\midrule
0.05 & 69.2 & 44.3 & 77.0 & 65.9 \\
\rowcolor{Periwinkle!10}
0.07 & \textbf{69.7} & \textbf{45.3} & \textbf{77.3} & \textbf{66.4} \\
0.20 & 68.9 & 43.7 & 76.7 & 65.5 \\
\bottomrule
\end{tabular}
\end{subtable}
\hfill
\begin{subtable}[t]{0.33\textwidth}
\centering
\caption{Alignment layer $m$}
\begin{tabular}{lcccc}
\toprule
$m$ & Image & Video & VisDoc & All \\
\midrule
14 & 68.8 & 42.4 & 74.9 & 64.6 \\
\rowcolor{Periwinkle!10}
21 & 69.7 & \textbf{45.3} & \textbf{77.3} & \textbf{66.4} \\
27 & \textbf{69.9} & 43.5 & 76.8 & 65.9 \\
\bottomrule
\end{tabular}
\end{subtable}
\end{table}

\paragraphnew{Additional InfoNCE training.}
Training InfoNCE for a second epoch raises the overall score by 0.2 points (63.7 $\rightarrow$ 63.9; \cref{tab:supp_longer_training}).
HEME reaches 66.4, with the largest gains on Video (+4.0) and VisDoc (+3.4) over the two-epoch baseline.

\begin{table}[H]
\centering
\caption{\textbf{Additional InfoNCE training on MMEB-V2.} All models use Qwen2-VL-2B.
InfoNCE is trained for one or two epochs; HEME trains the teacher and the student for one epoch each.
Bold marks the highest score in each column.}
\label{tab:supp_longer_training}
\footnotesize
\renewcommand{\arraystretch}{1.1}
\setlength{\tabcolsep}{3pt}
\begin{tabular}{lcccc}
\toprule
Method & Image & Video & VisDoc & All \\
\midrule
InfoNCE (1 epoch)  & 68.2 & 42.4 & 73.0 & 63.7 \\
InfoNCE (2 epochs) & 68.5 & 41.3 & 73.9 & 63.9 \\
\rowcolor{Periwinkle!10}
\textbf{HEME (Ours)} & \textbf{69.7} & \textbf{45.3} & \textbf{77.3} & \textbf{66.4} \\
\bottomrule
\end{tabular}
\end{table}

\clearpage

\subsection{Results by Dataset}
We report published task-level MMEB-V2 scores and category averages for standard embedding and explicit reasoning baselines. Scores for VLM2Vec-V2 and GME~\cite{zhang2025bridging} come from the benchmark report~\cite{meng2025vlm2vecv2}. For UME-R1~\cite{lan2026umer1}, TTE$_s$~\cite{cui2025think}, and RIME~\cite{wu2026beyond}, we use the scores from their respective papers.

\paragraphnew{Image tasks.}
\Cref{tab:supp_mmebv2_image} groups the 36 datasets into classification, question answering, retrieval, and grounding. HEME-2B and HEME-7B have mean Hit@1 scores of 69.7 and 73.9, respectively.

\definecolor{suppAvg}{RGB}{243,248,254}
\definecolor{suppCat}{RGB}{254,248,242}

\begin{table}[H]
\centering
\caption{\textbf{MMEB-V2 image results (Hit@1).} The first row gives the overall MMEB-V2 score across all 78 datasets. Bold marks the highest score in each row.}
\label{tab:supp_mmebv2_image}
\footnotesize
\setlength{\tabcolsep}{1pt}
\renewcommand{\arraystretch}{1.16}
\setlength{\aboverulesep}{1pt}
\setlength{\belowrulesep}{1pt}
\begin{tabularx}{\linewidth}{@{}lYYc*{8}{Y}@{}}
\toprule
Dataset & \multicolumn{2}{c}{GME} & VLM2Vec-V2 & \multicolumn{2}{c}{UME-R1} & \multicolumn{2}{c}{TTE$_s$} & \multicolumn{2}{c}{RIME} & \multicolumn{2}{c}{HEME} \\
\cmidrule(lr){2-3}\cmidrule(lr){5-6}\cmidrule(lr){7-8}\cmidrule(lr){9-10}\cmidrule(lr){11-12}
 & 2B & 7B & 2B & 2B & 7B & 2B & 7B & 2B & 7B & 2B & 7B \\
\midrule
\rowcolor{suppAvg} Avg -- All (78 tasks) & 54.1 & 57.8 & 58.0 & 60.1 & 64.5 & 63.1 & 68.6 & 64.1 & 68.6 & 66.4 & \textbf{70.4} \\
\rowcolor{suppAvg} Avg -- Image (36 tasks) & 51.9 & 56.0 & 64.9 & 66.6 & 71.3 & 70.1 & \textbf{74.2} & 69.1 & 73.4 & 69.7 & 73.9 \\
\midrule
\rowcolor{suppCat} CLS (10) & 54.4 & 57.7 & 62.9 & 64.8 & 67.1 & 67.9 & 69.7 & 67.9 & \textbf{70.3} & 66.3 & 69.5 \\
\rowcolor{suppCat} QA (10) & 29.9 & 34.7 & 56.3 & 62.8 & 69.2 & 66.6 & \textbf{72.4} & 64.4 & 71.7 & 66.5 & 71.8 \\
\rowcolor{suppCat} RET (12) & 66.9 & 71.2 & 69.5 & 67.6 & 71.9 & 70.2 & 74.0 & 69.8 & 73.2 & 71.4 & \textbf{74.9} \\
\rowcolor{suppCat} GD (4) & 55.5 & 59.3 & 77.3 & 77.2 & 84.9 & 84.1 & \textbf{90.6} & 82.1 & 86.3 & 81.6 & 87.0 \\
\midrule
ImageNet-1K & 58.3 & 64.6 & 80.8 & 75.3 & 80.4 & 83.3 & \textbf{84.3} & 81.2 & 80.9 & 81.0 & 83.0 \\
N24News & 50.1 & 50.5 & 72.9 & 81.1 & 82.3 & 78.6 & \textbf{83.1} & 80.0 & 82.7 & 78.3 & 82.8 \\
HatefulMemes & 52.5 & 53.6 & 56.3 & 75.2 & \textbf{79.0} & 64.0 & 67.4 & 68.4 & 76.2 & 64.2 & 71.7 \\
VOC2007 & 75.9 & 80.3 & 85.0 & 80.0 & 90.8 & 86.3 & 86.6 & 90.4 & \textbf{91.0} & 86.2 & 89.0 \\
SUN397 & 67.3 & 69.5 & 71.0 & 79.4 & 80.3 & 77.5 & 78.9 & 80.1 & 80.6 & 80.7 & \textbf{81.5} \\
Place365 & 35.8 & 39.1 & 35.9 & 42.6 & \textbf{46.8} & 45.7 & 44.6 & 45.3 & 45.5 & 44.1 & 46.0 \\
ImageNet-A & 28.8 & 41.2 & 47.4 & 50.4 & 53.9 & 50.9 & \textbf{60.4} & 52.1 & 57.4 & 50.9 & 54.5 \\
ImageNet-R & 78.6 & 83.9 & 89.3 & 88.7 & 90.1 & 89.7 & 90.5 & 89.9 & 89.9 & 90.9 & \textbf{91.2} \\
ObjectNet & 70.6 & 69.0 & 65.2 & 52.0 & 42.3 & \textbf{74.1} & 72.6 & 66.1 & 72.7 & 64.6 & 68.6 \\
Country211 & 26.5 & 24.8 & 25.2 & 23.4 & 25.0 & 28.5 & \textbf{29.0} & 25.5 & 26.4 & 22.4 & 26.7 \\
\midrule
OK-VQA & 29.9 & 33.2 & 51.5 & 62.4 & 71.7 & 68.4 & 74.7 & 65.8 & 74.1 & 68.5 & \textbf{74.8} \\
A-OKVQA & 18.6 & 21.0 & 43.6 & 51.1 & 58.7 & 57.1 & \textbf{66.1} & 56.4 & 61.8 & 58.6 & 62.9 \\
DocVQA & 29.8 & 41.4 & 90.1 & 92.2 & 93.8 & 94.2 & \textbf{95.6} & 93.4 & 94.4 & 94.3 & 95.0 \\
InfographicsVQA & 11.6 & 20.3 & 58.8 & 67.7 & \textbf{79.2} & 65.6 & 77.5 & 63.6 & 79.1 & 69.7 & 78.3 \\
ChartQA & 13.4 & 17.8 & 47.4 & 64.9 & 75.1 & 57.5 & 70.9 & 61.7 & \textbf{77.4} & 57.0 & 68.7 \\
Visual7W & 16.2 & 22.2 & 52.9 & 54.1 & 55.2 & 54.1 & 57.9 & 54.8 & 54.9 & 54.6 & \textbf{61.2} \\
ScienceQA & 27.3 & 28.0 & 38.2 & 42.7 & 53.7 & 50.7 & \textbf{60.0} & 45.7 & 59.0 & 50.0 & 59.1 \\
VizWiz & 37.0 & 39.0 & 43.3 & 46.8 & 51.6 & 55.1 & 53.8 & 47.7 & \textbf{55.3} & 52.7 & 55.1 \\
GQA & 75.1 & 76.9 & 64.9 & 67.3 & 69.3 & 77.0 & \textbf{80.9} & 73.1 & 73.6 & 75.3 & 75.6 \\
TextVQA & 39.7 & 46.8 & 72.2 & 78.6 & 83.5 & 86.2 & 87.0 & 81.3 & 87.0 & 84.5 & \textbf{87.5} \\
\midrule
VisDial & 48.1 & 60.8 & 82.7 & 76.6 & 80.7 & 81.2 & 84.4 & 80.7 & 82.6 & 83.4 & \textbf{85.9} \\
CIRR & 44.2 & 54.9 & 57.5 & 53.7 & 55.3 & 59.4 & \textbf{65.1} & 57.0 & 60.0 & 55.7 & 63.1 \\
VisualNews\_t2i & 74.7 & 79.7 & 74.5 & 71.7 & 76.8 & 72.8 & 78.5 & 72.4 & 79.8 & 75.4 & \textbf{80.7} \\
VisualNews\_i2t & 78.3 & \textbf{83.6} & 78.2 & 74.2 & 82.0 & 76.5 & 81.3 & 78.2 & 83.5 & 78.8 & \textbf{83.6} \\
MSCOCO\_t2i & 68.1 & 71.2 & 75.3 & 75.1 & 78.3 & 75.2 & 77.9 & 76.1 & 77.8 & 77.3 & \textbf{79.5} \\
MSCOCO\_i2t & 63.1 & 57.7 & 71.4 & 68.9 & 71.4 & 71.1 & 73.1 & 70.0 & 72.6 & 72.5 & \textbf{75.3} \\
NIGHTS & 67.0 & 67.6 & 68.6 & 67.2 & 68.1 & \textbf{70.8} & 69.8 & 67.9 & 68.6 & 69.1 & 69.3 \\
WebQA & 88.8 & 91.4 & 90.6 & 90.0 & 90.9 & 90.4 & 90.8 & 90.7 & 90.7 & 91.3 & \textbf{91.5} \\
FashionIQ & 32.9 & \textbf{37.8} & 19.5 & 17.1 & 23.4 & 26.3 & 29.7 & 19.8 & 26.0 & 21.5 & 27.4 \\
Wiki-SS-NQ & 73.9 & \textbf{78.2} & 66.9 & 62.0 & 72.5 & 64.2 & 70.5 & 68.9 & 76.3 & 72.2 & 77.1 \\
OVEN & 72.3 & \textbf{75.1} & 64.3 & 66.9 & 71.4 & 67.6 & 72.7 & 67.5 & 68.6 & 68.8 & 71.2 \\
EDIS & 91.8 & \textbf{96.0} & 84.1 & 88.0 & 92.0 & 87.0 & 93.9 & 88.9 & 91.6 & 90.3 & 93.7 \\
\midrule
MSCOCO & 28.6 & 31.4 & 67.1 & 69.5 & 72.7 & 67.7 & 74.1 & 69.0 & 72.0 & 68.6 & \textbf{74.2} \\
RefCOCO & 55.9 & 60.9 & 87.1 & 83.3 & 91.4 & 91.4 & \textbf{97.7} & 88.9 & 91.7 & 89.3 & 93.5 \\
RefCOCO-Matching & 73.3 & 78.4 & 85.8 & 84.4 & 91.1 & 95.0 & \textbf{96.3} & 90.1 & 93.7 & 90.5 & 93.9 \\
Visual7W-Pointing & 64.1 & 66.5 & 69.2 & 71.5 & 84.2 & 82.5 & \textbf{94.3} & 80.5 & 87.9 & 78.0 & 86.5 \\
\bottomrule
\end{tabularx}
\end{table}

\clearpage
\paragraphnew{Video tasks.}
\Cref{tab:supp_mmebv2_video} reports Hit@1 for classification, question answering, retrieval, and moment retrieval. Across the 18 datasets, HEME-2B averages 45.3 and HEME-7B averages 49.7.

\begin{table}[H]
\centering
\caption{\textbf{MMEB-V2 video results (Hit@1).} Bold marks the highest score in each row.}
\label{tab:supp_mmebv2_video}
\scriptsize
\setlength{\tabcolsep}{1pt}
\renewcommand{\arraystretch}{1.03}
\setlength{\aboverulesep}{1pt}
\setlength{\belowrulesep}{1pt}
\begin{tabularx}{\linewidth}{@{}lYYc*{8}{Y}@{}}
\toprule
Dataset & \multicolumn{2}{c}{GME} & VLM2Vec-V2 & \multicolumn{2}{c}{UME-R1} & \multicolumn{2}{c}{TTE$_s$} & \multicolumn{2}{c}{RIME} & \multicolumn{2}{c}{HEME} \\
\cmidrule(lr){2-3}\cmidrule(lr){5-6}\cmidrule(lr){7-8}\cmidrule(lr){9-10}\cmidrule(lr){11-12}
 & 2B & 7B & 2B & 2B & 7B & 2B & 7B & 2B & 7B & 2B & 7B \\
\midrule
\rowcolor{suppAvg} Avg -- Video (18 tasks) & 33.9 & 38.6 & 34.9 & 42.2 & 47.5 & 41.3 & 46.8 & 43.7 & 49.4 & 45.3 & \textbf{49.7} \\
\midrule
\rowcolor{suppCat} CLS (5) & 34.9 & 37.4 & 39.3 & 44.3 & 48.6 & 47.3 & 49.1 & 48.0 & 52.6 & 49.4 & \textbf{53.2} \\
\rowcolor{suppCat} QA (5) & 42.0 & 50.4 & 34.3 & 51.2 & 60.7 & 49.1 & 60.6 & 52.1 & \textbf{62.0} & 53.2 & 60.2 \\
\rowcolor{suppCat} RET (5) & 25.6 & 28.4 & 28.8 & 32.9 & 38.2 & 33.2 & 36.4 & 33.6 & 38.4 & 37.1 & \textbf{40.7} \\
\rowcolor{suppCat} MRET (3) & 32.4 & 38.2 & 38.5 & 39.7 & 39.3 & 32.1 & 37.2 & 39.2 & \textbf{41.6} & 39.2 & 41.5 \\
\midrule
K700 & 35.2 & 39.7 & 38.0 & 35.8 & 42.8 & 49.6 & \textbf{55.0} & 47.7 & \textbf{55.0} & 52.4 & 54.9 \\
SSv2 & 29.9 & 30.6 & 42.8 & 44.1 & 50.4 & 50.4 & 44.9 & 48.5 & 55.1 & 49.6 & \textbf{55.5} \\
HMDB51 & 43.4 & 47.9 & 40.9 & 54.4 & 58.3 & 52.5 & 51.7 & 56.7 & \textbf{58.9} & 54.6 & 54.3 \\
UCF101 & 52.4 & 54.7 & 60.0 & 67.2 & 70.0 & 58.3 & 64.2 & 66.4 & 68.5 & 69.5 & \textbf{70.3} \\
Breakfast & 13.6 & 14.3 & 14.8 & 20.1 & 21.5 & 25.4 & 29.7 & 20.6 & 25.4 & 20.8 & \textbf{30.9} \\
\midrule
MVBench & 37.5 & 46.6 & 33.7 & 49.9 & 58.2 & 48.5 & 59.5 & 49.1 & \textbf{59.9} & 46.9 & 53.5 \\
Video-MME & 34.3 & 39.2 & 30.7 & 41.7 & 47.3 & 45.8 & \textbf{53.1} & 41.6 & 49.6 & 42.4 & 48.4 \\
NExT-QA & 39.5 & 53.6 & 20.9 & 59.9 & 69.6 & 53.8 & \textbf{70.1} & 58.9 & 69.7 & 59.6 & 69.2 \\
EgoSchema & 40.8 & 46.8 & 34.0 & 45.4 & 52.4 & 36.4 & \textbf{55.6} & 40.4 & \textbf{55.6} & 41.0 & 53.4 \\
ActivityNetQA & 58.0 & 65.6 & 52.3 & 57.8 & 76.0 & 60.8 & 64.6 & 70.5 & 75.1 & 76.2 & \textbf{76.5} \\
\midrule
DiDeMo & 22.0 & 26.4 & 30.4 & 32.4 & 40.0 & 33.5 & 34.9 & 33.8 & 38.4 & 37.5 & \textbf{42.7} \\
MSR-VTT & 27.3 & 31.8 & 28.3 & 34.3 & 38.9 & 34.8 & 37.6 & 36.4 & 41.5 & 39.9 & \textbf{42.3} \\
MSVD & 47.6 & 49.7 & 48.1 & 55.4 & 60.8 & 56.5 & 58.5 & 56.6 & 59.4 & 60.9 & \textbf{63.3} \\
VATEX & 23.0 & 24.9 & 26.5 & 29.9 & 32.6 & 25.6 & 31.0 & 28.1 & 32.7 & 32.1 & \textbf{36.2} \\
YouCook2 & 7.9 & 9.1 & 10.6 & 12.7 & 18.5 & 15.8 & \textbf{19.9} & 13.3 & \textbf{19.9} & 15.1 & 19.0 \\
\midrule
QVHighlights & 43.6 & \textbf{59.5} & 49.4 & 57.5 & 54.9 & 38.9 & 51.0 & 55.1 & 56.9 & 47.8 & 53.1 \\
Charades-STA & 14.9 & 14.0 & 20.2 & 20.4 & 21.9 & 19.5 & 18.9 & 19.4 & 21.6 & 22.4 & \textbf{23.1} \\
MomentSeeker & 34.8 & 37.4 & 40.8 & 41.2 & 41.1 & 37.7 & 41.5 & 43.2 & 46.2 & 47.4 & \textbf{48.4} \\
\bottomrule
\end{tabularx}
\end{table}

\paragraphnew{Visual document tasks.}
\Cref{tab:supp_mmebv2_visdoc} reports NDCG@5 by dataset. The out-of-domain group includes separate page- and document-level results for ViDoSeek and MMLongBench-Doc.

\begin{table}[H]
\centering
\caption{\textbf{MMEB-V2 visual document results (NDCG@5).} Bold marks the highest score in each row.}
\label{tab:supp_mmebv2_visdoc}
\scriptsize
\setlength{\tabcolsep}{1pt}
\renewcommand{\arraystretch}{1.03}
\setlength{\aboverulesep}{1pt}
\setlength{\belowrulesep}{1pt}
\begin{tabularx}{\linewidth}{@{}lYYc*{8}{Y}@{}}
\toprule
Dataset & \multicolumn{2}{c}{GME} & VLM2Vec-V2 & \multicolumn{2}{c}{UME-R1} & \multicolumn{2}{c}{TTE$_s$} & \multicolumn{2}{c}{RIME} & \multicolumn{2}{c}{HEME} \\
\cmidrule(lr){2-3}\cmidrule(lr){5-6}\cmidrule(lr){7-8}\cmidrule(lr){9-10}\cmidrule(lr){11-12}
 & 2B & 7B & 2B & 2B & 7B & 2B & 7B & 2B & 7B & 2B & 7B \\
\midrule
\rowcolor{suppAvg} Avg -- VisDoc (24 tasks) & 72.7 & 75.2 & 65.4 & 63.9 & 67.1 & 68.8 & 76.4 & 71.4 & 75.6 & 77.3 & \textbf{80.6} \\
\midrule
\rowcolor{suppCat} ViDoRe-V1 (10) & 86.1 & \textbf{89.4} & 75.5 & 72.4 & 75.7 & 77.5 & 84.1 & 76.4 & 80.9 & 84.9 & 88.1 \\
\rowcolor{suppCat} ViDoRe-V2 (4) & 54.0 & 55.6 & 44.9 & 46.2 & 50.5 & 53.2 & \textbf{62.7} & 51.4 & 55.6 & 55.3 & 60.9 \\
\rowcolor{suppCat} VisRAG (6) & 82.5 & 85.0 & 79.4 & 79.2 & 83.7 & 83.2 & \textbf{91.9} & 81.7 & 85.8 & 87.3 & 88.7 \\
\rowcolor{suppCat} OOD (4) & 43.1 & 44.4 & 39.4 & 37.2 & 37.6 & 41.1 & 47.6 & 63.9 & 66.9 & 65.7 & \textbf{69.3} \\
\midrule
ViDoRe\_arxivqa & 82.8 & 86.9 & 80.6 & 73.9 & 73.6 & 80.7 & 84.6 & 82.1 & 84.1 & 89.7 & \textbf{91.4} \\
ViDoRe\_docvqa & 53.1 & 57.5 & 44.9 & 37.9 & 41.1 & 44.5 & 46.0 & 45.4 & 46.9 & 57.5 & \textbf{61.0} \\
ViDoRe\_infovqa & 90.2 & \textbf{91.6} & 83.7 & 76.2 & 80.8 & 84.8 & 88.7 & 81.2 & 85.4 & 89.9 & 91.3 \\
ViDoRe\_tabfquad & 93.3 & 94.6 & 89.2 & 86.1 & 90.2 & 88.4 & 94.7 & 85.5 & \textbf{95.3} & 90.7 & 94.1 \\
ViDoRe\_tatdqa & 69.9 & 74.1 & 43.8 & 40.6 & 46.7 & 50.4 & 59.4 & 48.9 & 52.6 & 69.8 & \textbf{75.4} \\
ViDoRe\_shiftproject & 89.5 & \textbf{96.8} & 60.8 & 66.8 & 65.0 & 65.2 & 81.6 & 69.9 & 73.7 & 77.7 & 85.3 \\
ViDoRe\_artificial\_intelligence & 97.5 & \textbf{99.6} & 88.5 & 85.9 & 89.5 & 91.9 & 98.1 & 89.8 & 96.1 & 96.8 & 96.9 \\
ViDoRe\_energy & 91.9 & \textbf{95.3} & 86.5 & 83.3 & 85.7 & 88.7 & 93.5 & 86.2 & 88.5 & 89.6 & 95.1 \\
ViDoRe\_government\_reports & 94.6 & \textbf{98.8} & 85.0 & 82.6 & 89.8 & 86.9 & 96.7 & 86.4 & 91.7 & 92.5 & 92.6 \\
ViDoRe\_healthcare\_industry & 98.7 & \textbf{99.3} & 92.2 & 90.8 & 94.3 & 92.8 & 97.9 & 88.3 & 94.9 & 94.3 & 98.4 \\
\midrule
ViDoRe\_esg\_reports\_human\_labeled\_v2 & 61.0 & 63.4 & 45.6 & 50.2 & 50.4 & 59.0 & \textbf{69.4} & 56.5 & 60.2 & 60.2 & 65.0 \\
ViDoRe\_biomedical\_lectures\_v2\_multilingual & 54.0 & 49.5 & 44.3 & 46.2 & 50.7 & 52.0 & \textbf{60.8} & 47.6 & 51.5 & 53.1 & 59.9 \\
ViDoRe\_economics\_reports\_v2\_multilingual & 50.2 & 54.2 & 43.0 & 45.7 & 57.8 & 49.8 & 60.4 & 47.0 & 59.2 & 53.4 & \textbf{62.0} \\
ViDoRe\_esg\_reports\_v2\_multilingual & 50.7 & 55.4 & 46.6 & 42.7 & 43.2 & 52.1 & \textbf{60.3} & 54.3 & 51.6 & 54.5 & 56.6 \\
\midrule
VisRAG\_ArxivQA & 82.0 & 87.4 & 76.9 & 74.3 & 80.5 & 78.5 & \textbf{94.5} & 79.4 & 84.0 & 85.7 & 89.1 \\
VisRAG\_ChartQA & 79.9 & 81.9 & 84.4 & 86.0 & 85.0 & 84.4 & \textbf{91.2} & 83.3 & 85.4 & 89.8 & 87.5 \\
VisRAG\_MP-DocVQA & 84.4 & 89.2 & 71.8 & 75.6 & 83.4 & 79.2 & 90.1 & 82.3 & 87.4 & 88.1 & \textbf{90.5} \\
VisRAG\_SlideVQA & 93.4 & 94.5 & 91.5 & 87.1 & 91.5 & 92.3 & \textbf{95.6} & 91.0 & 94.1 & 94.5 & \textbf{95.6} \\
VisRAG\_InfoVQA & 91.4 & 93.5 & 85.7 & 84.4 & 89.2 & 87.2 & 93.0 & 82.4 & 91.8 & 90.9 & \textbf{93.9} \\
VisRAG\_PlotQA & 64.1 & 63.4 & 66.1 & 68.0 & 72.7 & 77.5 & \textbf{86.9} & 71.6 & 72.1 & 74.6 & 75.4 \\
\midrule
ViDoSeek-page & 21.6 & 23.2 & 21.9 & 21.2 & 21.3 & 22.6 & 35.0 & 80.7 & \textbf{85.6} & 79.1 & 82.6 \\
ViDoSeek-doc & 83.6 & 83.9 & 80.2 & 75.9 & 75.3 & 82.0 & \textbf{84.4} & 79.7 & 80.9 & 82.6 & 83.9 \\
MMLongBench-page & 15.8 & 16.2 & 11.9 & 11.9 & 12.3 & 12.9 & 20.7 & 48.3 & 52.4 & 50.1 & \textbf{55.9} \\
MMLongBench-doc & 51.4 & 54.3 & 43.7 & 39.7 & 41.3 & 47.0 & 50.4 & 46.9 & 48.8 & 50.9 & \textbf{54.7} \\
\bottomrule
\end{tabularx}
\end{table}

\clearpage

\section{MMEB-V3 Evaluation}
\label{sec:supp_mmebv3}
\paragraphnew{Benchmark and metrics.}
We follow the MMEB-V3 evaluation protocol~\cite{huang2026mmebv3} and average scores over datasets.
\Cref{tab:supp_mmebv3} lists the 47 agent and 53 text tasks used for evaluation.
Each query ranks the full candidate corpus for its task. 
Inputs longer than Qwen2-VL's context length of 32,768 tokens are truncated at the end of the text, keeping the readout token.
We report Hit@1 for agent and NDCG@5 for text tasks.
These benchmark datasets are not included in our HEME fine-tuning mixture. We omit the 11 audio tasks because Qwen2-VL does not accept audio input.

\begin{table}[H]
\centering
\caption{\textbf{MMEB-V3 agent and text tasks.} Parentheses give the number of tasks from each source.}
\label{tab:supp_mmebv3}
\small
\renewcommand{\arraystretch}{1.13}
\setlength{\tabcolsep}{4pt}
\begin{tabularx}{\linewidth}{@{}llXr@{}}
\toprule
\textbf{Category} & \textbf{Sub-category} & \textbf{Sources} & \textbf{\#} \\
\midrule
Agent & Tool retrieval & Tool-REX (35)  & 35 \\
 & GUI retrieval & GAE-Bench: GUIAct (4), Mind2Web (4) & 8 \\
 & Memory retrieval & KnowMeBench, REALTALK, PeerQA, DeepPlanning & 4 \\
\midrule
Text & Reasoning-intensive retrieval (RR) & BRIGHT (12), R2MED (8) & 20 \\
 & Instruction-following retrieval (IF) & FollowIR (3), InfoSearch (6) & 9 \\
 & Long-context retrieval (LC) & LongEmbed (6) & 6 \\
 & Multi-condition retrieval (MC) & MultiConIR (5) & 5 \\
 & General retrieval (GR) & NanoBEIR (13) & 13 \\
\midrule
\multicolumn{3}{@{}l}{Total} & 100 \\
\bottomrule
\end{tabularx}
\end{table}

\paragraphnew{Subcategory results.}
Across the 100 tasks in \cref{tab:supp_mmebv3_summary}, HEME-2B averages 35.6 and HEME-7B averages 38.7. Baseline scores and their subcategory averages are from the benchmark report~\cite{huang2026mmebv3}. For models evaluated on audio in that report, we recompute the overall average using only the agent and text tasks.

\begin{table}[H]
\centering
\caption{\textbf{MMEB-V3 results by subcategory.} Agent scores use Hit@1 and text scores use NDCG@5. The last column averages all 100 tasks. $^\dagger$Models use additional text-only training data.}
\label{tab:supp_mmebv3_summary}
\footnotesize
\renewcommand{\arraystretch}{1.13}
\setlength{\tabcolsep}{2pt}
\begin{tabularx}{\linewidth}{@{}llYYc*{8}{Y}@{}}
\toprule
\multirow{2}{*}{\textbf{Model}} & \multirow{2}{*}{\textbf{Backbone}} & \multicolumn{4}{c}{\textbf{Agent}} & \multicolumn{6}{c}{\textbf{Text}} & \multirow{2}{*}{\textbf{All}} \\
\cmidrule(lr){3-6}\cmidrule(lr){7-12}
 & & Tool & GUI & Memory & Avg & RR & IF & LC & MC & GR & Avg & \\
\midrule
\# Tasks & & 35 & 8 & 4 & 47 & 20 & 9 & 6 & 5 & 13 & 53 & 100 \\
\midrule
VLM2Vec-V2~\cite{meng2025vlm2vecv2} & Qwen2-VL-2B & 27.6 & 36.2 & 23.3 & 28.7 & 7.8 & 29.2 & 11.5 & 50.3 & 41.2 & 24.5 & 26.5 \\
E5-Omni$^{\dagger}$~\cite{chen2026e5omni} & Qwen2.5-Omni-3B & 37.7 & 35.6 & 32.3 & 36.9 & 11.5 & 44.4 & 8.3 & 60.7 & 34.4 & 26.7 & 31.5 \\
Omni-Embed-Nemotron$^{\dagger}$~\cite{xu2025omniembednemotron} & Qwen2.5-Omni-3B & 38.1 & 32.0 & 32.2 & 36.5 & 17.2 & 42.8 & 50.0 & 69.7 & 56.9 & 39.2 & 37.9 \\
Qwen3-VL-Embed$^{\dagger}$~\cite{li2026qwen3vlembedding} & Qwen3-VL-2B & 42.6 & 30.4 & 28.4 & 39.3 & 16.6 & 40.6 & 53.1 & 61.2 & 58.0 & 39.2 & 39.2 \\
\rowcolor{Periwinkle!10}
\textbf{HEME (Ours)} & Qwen2-VL-2B & 37.6 & 39.5 & 24.8 & 36.9 & 11.7 & 39.0 & 48.8 & 59.0 & 50.1 & 34.4 & 35.6 \\
\midrule
VLM2Vec-Qwen2VL~\cite{jiang2024vlm2vec} & Qwen2-VL-7B & 19.8 & 21.4 & 15.9 & 19.7 & 7.2 & 28.1 & 5.9 & 40.9 & 41.7 & 22.2 & 21.0 \\
LCO-Embed-Omni$^{\dagger}$~\cite{xiao2025scaling} & Qwen2.5-Omni-7B & 29.0 & 25.0 & 23.0 & 27.8 & 13.6 & 52.0 & 12.6 & 61.4 & 52.1 & 32.4 & 30.2 \\
E5-Omni$^{\dagger}$~\cite{chen2026e5omni} & Qwen2.5-Omni-7B & 37.4 & 38.0 & 27.3 & 36.7 & 11.1 & 42.1 & 9.7 & 56.9 & 35.9 & 26.9 & 31.5 \\
GME$^{\dagger}$~\cite{zhang2025bridging} & Qwen2-VL-7B & 39.0 & 30.0 & 17.1 & 35.6 & 12.5 & 52.4 & 17.8 & 59.0 & 62.5 & 37.1 & 36.4 \\
Qwen3-VL-Embed$^{\dagger}$~\cite{li2026qwen3vlembedding} & Qwen3-VL-8B & 41.3 & 33.5 & 22.8 & 38.4 & 18.2 & 44.8 & 58.0 & 61.2 & 61.2 & 42.5 & 40.6 \\
\rowcolor{Periwinkle!10}
\textbf{HEME (Ours)} & Qwen2-VL-7B & 40.4 & 43.1 & 27.9 & 39.8 & 15.7 & 42.7 & 51.5 & 57.6 & 53.9 & 37.7 & 38.7 \\
\bottomrule
\end{tabularx}
\end{table}

\clearpage
\subsection{Results by Task}
\paragraphnew{Agent tasks.}
\Cref{tab:supp_mmebv3_agent} lists the individual tool, GUI, and memory retrieval tasks. HEME-2B and HEME-7B average 36.9 and 39.8 Hit@1 across these tasks.

\begin{table}[H]
\centering
\caption{\textbf{MMEB-V3 agent results (Hit@1).} The first row includes both agent and text tasks. Bold marks the highest score in each row.}
\label{tab:supp_mmebv3_agent}
\footnotesize
\setlength{\tabcolsep}{1pt}
\renewcommand{\arraystretch}{1.03}
\setlength{\aboverulesep}{1pt}
\setlength{\belowrulesep}{1pt}
\begin{tabularx}{\linewidth}{@{}lccYYcc*{5}{Y}@{}}
\toprule
Task & \shortstack{VLM2Vec\\-V2} & \shortstack{VLM2Vec\\-Qwen2VL} & \multicolumn{2}{c}{E5-Omni} & \shortstack{Omni-Embed\\-Nemotron} & \shortstack{LCO-Embed\\-Omni} & GME & \multicolumn{2}{c}{\shortstack{Qwen3-VL\\-Embedding}} & \multicolumn{2}{c}{HEME} \\
\cmidrule(lr){4-5}\cmidrule(lr){9-10}\cmidrule(lr){11-12}
 & 2B & 7B & 3B & 7B & 3B & 7B & 7B & 2B & 8B & 2B & 7B \\
\midrule
\rowcolor{suppAvg} Avg -- All (100 tasks) & 26.5 & 21.0 & 31.5 & 31.5 & 37.9 & 30.2 & 36.4 & 39.2 & \textbf{40.6} & 35.6 & 38.7 \\
\rowcolor{suppAvg} Avg -- Agent (47 tasks) & 28.7 & 19.7 & 36.9 & 36.7 & 36.5 & 27.8 & 35.6 & 39.3 & 38.4 & 36.9 & \textbf{39.8} \\
\midrule
\rowcolor{suppCat} Tool (35) & 27.6 & 19.8 & 37.7 & 37.4 & 38.1 & 29.0 & 39.0 & \textbf{42.6} & 41.3 & 37.6 & 40.4 \\
\rowcolor{suppCat} GUI (8) & 36.2 & 21.4 & 35.6 & 38.0 & 32.0 & 25.0 & 30.0 & 30.4 & 33.5 & 39.5 & \textbf{43.1} \\
\rowcolor{suppCat} Memory (4) & 23.3 & 15.9 & \textbf{32.3} & 27.3 & 32.2 & 23.0 & 17.1 & 28.4 & 22.8 & 24.8 & 27.9 \\
\midrule
Tool-craft-math-algebra & 65.0 & 57.1 & 75.7 & 81.4 & \textbf{85.0} & 63.9 & 84.6 & 84.6 & 83.9 & 74.3 & 80.0 \\
Tool-craft-tabmwp & 14.4 & 10.9 & 19.0 & 29.9 & 29.3 & 25.3 & 27.6 & \textbf{36.8} & 34.5 & 19.0 & 23.0 \\
Tool-craft-vqa & 43.5 & 23.0 & 47.0 & 50.5 & \textbf{65.5} & 34.5 & 60.5 & 52.0 & 54.5 & 46.0 & 51.0 \\
Tool-gorilla-huggingface & 20.2 & 18.6 & 25.8 & 19.2 & 26.6 & 20.2 & 24.6 & 26.4 & \textbf{28.4} & 25.0 & 27.4 \\
Tool-gorilla-pytorch & 4.7 & 0.0 & \textbf{20.9} & 14.0 & 11.6 & 16.3 & 14.0 & 11.6 & 9.3 & 9.3 & 14.0 \\
Tool-gorilla-tensor & 7.3 & 1.8 & 16.4 & 12.7 & 20.0 & 5.5 & \textbf{21.8} & 16.4 & 16.4 & 0.0 & 3.6 \\
Tool-toolink & 40.4 & 36.4 & 68.4 & \textbf{70.6} & 67.0 & 65.6 & 66.4 & 67.8 & 67.6 & 58.1 & 55.9 \\
Tool-apibank & 38.6 & 1.0 & 53.5 & 47.5 & 45.5 & 7.9 & 52.5 & \textbf{62.4} & 58.4 & 50.5 & 50.5 \\
Tool-apigen & 46.5 & 49.3 & 59.7 & 61.2 & \textbf{64.0} & 50.2 & 54.0 & 55.4 & 55.5 & 55.0 & 51.8 \\
Tool-mnms & 30.3 & 27.3 & \textbf{54.5} & 51.5 & 42.4 & 27.3 & 30.3 & 36.4 & 30.3 & 33.3 & \textbf{54.5} \\
Tool-reversechain & 51.5 & 22.5 & \textbf{64.0} & 56.5 & 50.0 & 18.5 & 54.5 & 61.0 & 57.5 & 56.5 & 59.5 \\
Tool-rotbench & 1.8 & 1.1 & 5.1 & 4.2 & 6.9 & 2.5 & \textbf{7.3} & 3.3 & 4.2 & 3.1 & 5.5 \\
Tool-t-eval-dialog & 22.0 & 26.0 & 16.0 & 24.0 & 14.0 & 20.0 & \textbf{34.0} & 24.0 & 20.0 & 18.0 & 30.0 \\
Tool-t-eval-step & 10.0 & \textbf{38.0} & 26.0 & 12.0 & 12.0 & 22.0 & 30.0 & 16.0 & 12.0 & 24.0 & 26.0 \\
Tool-taskbench-daily & 52.5 & 45.0 & \textbf{90.0} & 75.0 & 67.5 & 62.5 & 72.5 & 65.0 & 62.5 & 52.5 & 85.0 \\
Tool-toolace & 56.4 & 33.3 & 67.1 & \textbf{71.4} & 67.0 & 51.7 & 69.0 & 65.7 & 62.3 & 60.9 & 67.0 \\
Tool-toolbench & 24.8 & 6.0 & 29.2 & 28.9 & 13.3 & 16.9 & 17.4 & \textbf{40.6} & 36.0 & 28.6 & 35.5 \\
Tool-toolemu & 10.5 & 0.0 & 10.5 & 5.3 & \textbf{34.2} & 7.9 & 13.2 & 31.6 & 26.3 & 31.6 & 26.3 \\
Tool-tooleyes & 15.8 & 4.2 & \textbf{24.2} & 12.6 & 10.5 & 11.6 & 20.0 & 19.0 & 20.0 & 18.9 & 23.2 \\
Tool-toollens & 8.9 & 1.3 & 5.1 & 10.2 & 1.0 & 2.9 & 1.3 & 8.0 & 7.0 & 9.9 & \textbf{12.1} \\
Tool-ultratool & 4.6 & 0.4 & 44.0 & 43.8 & 43.6 & 41.2 & 40.4 & \textbf{50.6} & 41.2 & 30.2 & 42.2 \\
Tool-autotools-food & 4.6 & 9.1 & 9.1 & 18.2 & 13.6 & 4.5 & 18.2 & 50.0 & 50.0 & \textbf{54.5} & 40.9 \\
Tool-autotools-music & 0.0 & 0.0 & 0.0 & 0.0 & 3.1 & 0.0 & 0.0 & 12.5 & 37.5 & \textbf{59.4} & 28.1 \\
Tool-autotools-weather & 0.0 & 0.0 & 0.0 & 9.1 & 0.0 & 0.0 & 0.0 & 18.2 & 9.1 & \textbf{45.5} & 9.1 \\
Tool-restgpt-spotify & 12.5 & 27.5 & 20.0 & 30.0 & 25.0 & 0.0 & 20.0 & 60.0 & \textbf{67.5} & 7.5 & 20.0 \\
Tool-restgpt-tmdb & 0.0 & 0.0 & 9.3 & 13.0 & 16.7 & 9.3 & 11.1 & 37.0 & \textbf{38.9} & 13.0 & 11.1 \\
Tool-appbench & 65.6 & 65.6 & 87.5 & 84.4 & 87.5 & 59.4 & 81.2 & 71.9 & 84.4 & \textbf{90.6} & \textbf{90.6} \\
Tool-gpt4tools & 46.9 & 28.1 & 37.5 & 50.0 & 75.0 & 81.2 & 81.2 & \textbf{84.4} & 68.8 & 50.0 & 71.9 \\
Tool-gta & 0.0 & 14.3 & 28.6 & 14.3 & \textbf{42.9} & 28.6 & 35.7 & 14.3 & 21.4 & 14.3 & 21.4 \\
Tool-taskbench-huggingface & 56.5 & 4.3 & 47.8 & 56.5 & 52.2 & 52.2 & 56.5 & \textbf{60.9} & 39.1 & 47.8 & 47.8 \\
Tool-taskbench-multimedia & 65.0 & 32.5 & \textbf{92.5} & 82.5 & 77.5 & 60.0 & 87.5 & 70.0 & 62.5 & 72.5 & 80.0 \\
Tool-metatool & 44.0 & 32.0 & 46.0 & 51.0 & 49.0 & 48.5 & \textbf{54.0} & 52.5 & 51.5 & 52.5 & 51.0 \\
Tool-tool-be-honest & 46.0 & 47.1 & 54.0 & 52.0 & 54.3 & 35.4 & \textbf{55.4} & 51.1 & 49.1 & 43.7 & 50.3 \\
Tool-toolalpaca & 51.1 & 26.6 & 58.5 & 60.6 & 50.0 & 60.6 & 59.6 & 62.8 & \textbf{67.0} & 55.3 & 60.6 \\
Tool-toolbench-sam & 4.1 & 1.0 & 5.6 & 6.6 & 8.1 & 2.5 & 7.6 & \textbf{11.7} & \textbf{11.7} & 5.6 & 6.1 \\
\midrule
GAE-guiact-q2s & 25.9 & 7.7 & 29.9 & 32.3 & 26.1 & 24.1 & 31.3 & 19.8 & 29.5 & 31.8 & \textbf{37.2} \\
GAE-guiact-q2t & 52.6 & 42.6 & 50.9 & 55.5 & 39.8 & 42.8 & 31.3 & 48.4 & 46.3 & 52.6 & \textbf{57.4} \\
GAE-guiact-s2s & 35.8 & 33.2 & 36.5 & 36.4 & 35.9 & 17.9 & 32.7 & 25.5 & 30.2 & \textbf{44.6} & 43.1 \\
GAE-guiact-t2s & 24.1 & 8.6 & 21.9 & 23.0 & 20.7 & 13.8 & 17.1 & 25.0 & 24.9 & 33.0 & \textbf{33.6} \\
GAE-mind2web-q2s & 43.0 & 12.2 & 43.0 & 44.9 & 41.7 & 33.0 & 42.1 & 27.1 & 37.7 & 44.5 & \textbf{54.5} \\
GAE-mind2web-q2t & 47.8 & 32.4 & 40.3 & 50.2 & 33.6 & 25.3 & 32.4 & 44.3 & 37.2 & 46.6 & \textbf{51.4} \\
GAE-mind2web-s2s & 37.1 & 32.4 & 38.3 & 37.7 & 37.1 & 25.9 & 37.1 & 30.2 & 35.2 & \textbf{38.6} & \textbf{38.6} \\
GAE-mind2web-t2s & 23.4 & 1.9 & 23.7 & 23.7 & 21.5 & 17.1 & 15.9 & 23.1 & 27.4 & 24.0 & \textbf{28.7} \\
\midrule
KnowMeBench & 32.6 & 22.3 & \textbf{45.1} & 38.4 & 44.4 & 35.8 & 19.0 & 41.5 & 35.3 & 35.6 & 40.4 \\
REALTALK & 18.4 & 11.8 & 27.8 & 27.5 & 32.0 & 28.1 & 19.6 & \textbf{35.4} & 30.5 & 26.7 & 29.6 \\
PeerQA & 21.3 & 11.0 & \textbf{22.8} & 19.9 & 18.4 & 15.4 & 14.7 & 17.7 & 16.2 & 16.9 & 19.9 \\
DeepPlanning & 20.8 & 18.3 & 33.3 & 23.3 & \textbf{34.2} & 12.5 & 15.0 & 19.2 & 9.2 & 20.0 & 21.7 \\
\bottomrule
\end{tabularx}
\end{table}

\clearpage
\paragraphnew{Text tasks.}
\Cref{tab:supp_mmebv3_text} reports NDCG@5 on the 53 text retrieval tasks. HEME-2B and HEME-7B average 34.4 and 37.7, respectively. The reasoning-intensive group includes the 12 BRIGHT tasks and the eight R2MED tasks.

\begin{table}[H]
\centering
\caption{\textbf{MMEB-V3 text results (NDCG@5).} Tasks are grouped by retrieval subcategory. Bold marks the highest score in each row.}
\label{tab:supp_mmebv3_text}
\scriptsize
\setlength{\tabcolsep}{1pt}
\renewcommand{\arraystretch}{1.06}
\setlength{\aboverulesep}{1pt}
\setlength{\belowrulesep}{1pt}
\begin{tabularx}{\linewidth}{@{}lccYYcc*{5}{Y}@{}}
\toprule
Task & \shortstack{VLM2Vec\\-V2} & \shortstack{VLM2Vec\\-Qwen2VL} & \multicolumn{2}{c}{E5-Omni} & \shortstack{Omni-Embed\\-Nemotron} & \shortstack{LCO-Embed\\-Omni} & GME & \multicolumn{2}{c}{\shortstack{Qwen3-VL\\-Embedding}} & \multicolumn{2}{c}{HEME} \\
\cmidrule(lr){4-5}\cmidrule(lr){9-10}\cmidrule(lr){11-12}
 & 2B & 7B & 3B & 7B & 3B & 7B & 7B & 2B & 8B & 2B & 7B \\
\midrule
\rowcolor{suppAvg} Avg -- Text (53 tasks) & 24.5 & 22.2 & 26.7 & 26.9 & 39.2 & 32.4 & 37.1 & 39.2 & \textbf{42.5} & 34.4 & 37.7 \\
\midrule
\rowcolor{suppCat} RR (20) & 7.8 & 7.2 & 11.5 & 11.1 & 17.2 & 13.6 & 12.5 & 16.6 & \textbf{18.2} & 11.7 & 15.7 \\
\rowcolor{suppCat} IF (9) & 29.2 & 28.1 & 44.4 & 42.1 & 42.8 & 52.0 & \textbf{52.4} & 40.6 & 44.8 & 39.0 & 42.7 \\
\rowcolor{suppCat} LC (6) & 11.5 & 5.9 & 8.3 & 9.7 & 50.0 & 12.6 & 17.8 & 53.1 & \textbf{58.0} & 48.8 & 51.5 \\
\rowcolor{suppCat} MC (5) & 50.3 & 40.9 & 60.7 & 56.9 & \textbf{69.7} & 61.4 & 59.0 & 61.2 & 61.2 & 59.0 & 57.6 \\
\rowcolor{suppCat} GR (13) & 41.2 & 41.7 & 34.4 & 35.9 & 56.9 & 52.1 & \textbf{62.5} & 58.0 & 61.2 & 50.1 & 53.9 \\
\midrule
BRIGHT aops & 1.2 & 2.2 & 3.7 & 3.9 & 3.3 & \textbf{8.2} & 5.0 & 2.4 & 1.1 & 2.5 & 4.0 \\
BRIGHT biology & 12.3 & 9.4 & 2.0 & 1.6 & 13.6 & 11.1 & 3.0 & 14.7 & \textbf{17.1} & 7.7 & 11.8 \\
BRIGHT earth science & 11.4 & 0.7 & 7.7 & 6.1 & 26.4 & 13.6 & 12.8 & 26.2 & \textbf{28.9} & 14.5 & 15.6 \\
BRIGHT economics & 3.9 & 2.8 & 4.5 & 3.8 & 15.4 & 10.3 & 7.0 & 15.9 & \textbf{16.7} & 8.7 & 11.0 \\
BRIGHT leetcode & 8.5 & 4.7 & 17.9 & \textbf{20.3} & 18.5 & 5.6 & 17.3 & 13.7 & 17.3 & 13.4 & 15.6 \\
BRIGHT pony & 1.5 & \textbf{9.5} & 2.4 & 1.3 & 2.6 & 0.5 & 2.1 & 2.5 & 9.1 & 3.4 & 4.2 \\
BRIGHT psychology & 4.4 & 3.0 & 5.6 & 4.2 & 18.1 & 8.5 & 11.0 & \textbf{18.6} & 12.0 & 9.0 & 15.5 \\
BRIGHT robotics & 4.8 & 1.5 & 7.9 & 7.0 & 12.1 & 2.8 & 5.7 & 11.3 & \textbf{12.5} & 7.9 & 6.6 \\
BRIGHT stackoverflow & 2.1 & 0.0 & 6.8 & 6.0 & 9.6 & 1.6 & 6.1 & \textbf{14.5} & 12.0 & 4.9 & 7.5 \\
BRIGHT sustainable living & 2.6 & 3.6 & 4.7 & 4.9 & 6.6 & 10.9 & 9.7 & 9.7 & \textbf{14.2} & 6.2 & 9.0 \\
BRIGHT theoremqa questions & 9.4 & 15.1 & 12.3 & \textbf{16.9} & 13.9 & 9.9 & 12.8 & 10.6 & 12.8 & 11.9 & 14.2 \\
BRIGHT theoremqa theorems & 4.7 & 3.2 & 11.2 & 11.6 & \textbf{23.0} & 8.6 & 17.0 & 12.8 & 21.2 & 5.9 & 12.0 \\
\midrule
R2MED Bioinformatics & 19.4 & 15.1 & 17.5 & 33.7 & 36.9 & 14.0 & 33.2 & 35.8 & \textbf{42.2} & 23.0 & 34.6 \\
R2MED Biology & 12.3 & 9.3 & 2.0 & 1.6 & 13.9 & 11.0 & 3.2 & 14.7 & \textbf{17.2} & 7.6 & 11.9 \\
R2MED IIYi-Clinical & 8.5 & 2.3 & 15.5 & 21.9 & 15.7 & 4.2 & 18.5 & 21.3 & \textbf{26.3} & 16.2 & 18.3 \\
R2MED MedQA-Diag & 21.8 & 16.4 & 8.4 & 6.8 & 6.9 & 6.7 & 28.9 & 33.6 & \textbf{34.4} & 7.6 & 12.6 \\
R2MED MedXpertQA-Exam & 5.6 & 6.0 & 5.3 & 6.9 & 5.3 & 6.5 & 5.3 & 11.5 & \textbf{17.4} & 4.9 & 10.9 \\
R2MED Medical-Sciences & 4.7 & 3.5 & 25.8 & 20.0 & 27.3 & \textbf{33.1} & 9.3 & 10.1 & 13.1 & 27.0 & 29.2 \\
R2MED PMC-Clinical & 14.0 & 13.5 & 28.8 & 25.7 & 21.7 & 8.7 & 32.2 & 21.4 & \textbf{37.1} & 22.4 & 31.3 \\
R2MED PMC-Treatment & 28.3 & 21.4 & 27.6 & 34.9 & 14.2 & 13.9 & \textbf{42.0} & 30.4 & 37.4 & 30.5 & 39.0 \\
\midrule
FollowIR core17 & 21.3 & 21.0 & 28.6 & 26.9 & 30.1 & \textbf{45.3} & 34.8 & 34.5 & 34.3 & 41.1 & 44.3 \\
FollowIR news21 & 4.6 & 9.7 & 25.3 & 24.9 & 31.6 & 26.7 & 13.3 & 21.8 & 20.4 & 40.9 & \textbf{48.1} \\
FollowIR robust04 & 24.7 & 21.1 & 32.1 & 27.4 & 33.5 & 40.2 & 30.6 & 39.5 & \textbf{44.0} & 37.6 & 39.6 \\
InfoSearch Audience-v1 & 17.2 & 25.0 & 29.1 & 34.2 & 27.4 & 48.6 & \textbf{61.8} & 31.3 & 34.5 & 17.5 & 20.4 \\
InfoSearch Clarity-v1 & 43.6 & 34.5 & 62.3 & 43.8 & 34.5 & 50.5 & 57.9 & \textbf{62.8} & 55.4 & 36.7 & 28.5 \\
InfoSearch Format-v1 & 20.4 & 10.2 & 49.9 & 49.2 & \textbf{52.5} & 47.3 & 47.5 & 38.2 & 46.2 & 29.7 & 42.0 \\
InfoSearch Language-v1 & 49.2 & 45.3 & 67.9 & 66.5 & 71.3 & 68.9 & \textbf{77.5} & 56.5 & 67.6 & 57.2 & 67.1 \\
InfoSearch Length-v1 & 43.9 & 40.0 & 56.4 & 58.6 & 56.9 & 73.3 & \textbf{75.4} & 44.2 & 55.9 & 44.9 & 47.2 \\
InfoSearch Source-v1 & 37.6 & 46.1 & 47.8 & 46.9 & 47.7 & 67.0 & \textbf{72.5} & 36.8 & 44.5 & 45.7 & 47.2 \\
\midrule
LongEmbed 2wikimqa & 13.7 & 6.0 & 1.4 & 1.2 & 71.1 & 2.5 & 14.7 & 83.8 & \textbf{89.0} & 54.0 & 50.7 \\
LongEmbed narrativeqa & 3.6 & 1.3 & 1.2 & 1.1 & 32.7 & 1.8 & 6.8 & 52.9 & \textbf{60.9} & 45.5 & 55.6 \\
LongEmbed needle & 7.6 & 3.2 & 14.9 & 21.7 & 24.3 & 17.1 & 21.2 & 15.5 & 21.2 & 25.9 & \textbf{27.3} \\
LongEmbed passkey & 11.3 & 12.7 & 27.7 & 31.4 & 35.9 & 31.8 & 26.7 & 32.8 & \textbf{38.2} & 31.0 & 34.1 \\
LongEmbed qmsum & 11.9 & 4.7 & 1.5 & 1.5 & 40.2 & 8.4 & 9.8 & 37.5 & 42.3 & 44.0 & \textbf{46.2} \\
LongEmbed summ screen fd & 20.9 & 7.5 & 3.1 & 1.3 & 96.0 & 13.8 & 27.4 & 95.9 & \textbf{96.3} & 92.4 & 95.0 \\
\midrule
MultiConIR Books & 47.1 & 35.0 & 52.1 & 52.5 & \textbf{64.0} & 56.1 & 56.8 & 62.1 & 60.9 & 55.8 & 55.4 \\
MultiConIR Legal Document & 52.4 & 38.0 & 57.7 & 54.7 & \textbf{68.8} & 59.8 & 56.7 & 57.2 & 57.7 & 56.7 & 60.8 \\
MultiConIR Medical Case & 43.9 & 35.8 & 57.7 & 51.6 & \textbf{65.7} & 59.4 & 52.8 & 60.0 & 61.4 & 55.0 & 52.7 \\
MultiConIR Movies & 48.7 & 42.0 & 66.3 & 61.9 & \textbf{74.7} & 65.2 & 64.6 & 64.9 & 62.9 & 66.5 & 60.7 \\
MultiConIR People & 59.2 & 53.4 & 69.8 & 64.0 & \textbf{75.1} & 66.2 & 64.1 & 61.8 & 62.9 & 60.8 & 58.2 \\
\midrule
NanoBEIR NanoArguAna & 36.1 & 32.4 & 53.3 & 49.9 & 42.8 & 25.7 & \textbf{54.3} & 49.4 & 45.0 & 41.9 & 49.7 \\
NanoBEIR NanoClimateFEVER & 14.4 & 14.2 & 2.2 & 1.3 & 23.6 & 25.0 & 31.9 & 29.5 & \textbf{33.5} & 21.6 & 25.0 \\
NanoBEIR NanoDBPedia & 48.4 & 51.7 & 38.2 & 36.2 & 53.8 & \textbf{65.8} & 60.5 & 60.6 & 61.1 & 55.0 & 59.2 \\
NanoBEIR NanoFEVER & 52.3 & 55.9 & 18.6 & 19.0 & 75.4 & 80.0 & 87.1 & 89.2 & \textbf{89.8} & 77.4 & 76.4 \\
NanoBEIR NanoFiQA2018 & 21.0 & 20.1 & 15.4 & 24.0 & 54.4 & 43.2 & \textbf{58.8} & 43.7 & 53.2 & 37.4 & 39.3 \\
NanoBEIR NanoHotpotQA & 50.0 & 53.6 & 73.0 & 56.9 & \textbf{92.0} & 62.2 & 84.5 & 73.8 & 77.4 & 62.3 & 63.8 \\
NanoBEIR NanoMSMARCO & 38.6 & 28.1 & 47.0 & 44.4 & 53.4 & 45.7 & \textbf{65.5} & 53.6 & 64.9 & 46.1 & 48.0 \\
NanoBEIR NanoNFCorpus & 29.5 & 31.8 & 10.8 & 13.2 & 31.8 & 34.8 & \textbf{39.9} & 39.5 & \textbf{39.9} & 28.2 & 34.5 \\
NanoBEIR NanoNQ & 28.2 & 34.5 & 28.8 & 23.9 & 64.5 & 58.5 & 65.9 & 56.0 & \textbf{66.6} & 53.2 & 60.1 \\
NanoBEIR NanoQuoraRetrieval & 92.8 & 93.4 & 92.9 & 87.5 & 91.5 & 90.5 & 95.2 & 94.9 & 96.0 & 92.4 & \textbf{96.1} \\
NanoBEIR NanoSCIDOCS & 24.3 & 30.1 & 7.2 & 33.1 & 32.7 & 33.0 & \textbf{44.1} & 35.7 & 37.5 & 29.5 & 34.5 \\
NanoBEIR NanoSciFact & 58.4 & 61.7 & 33.5 & 43.4 & 73.8 & 66.9 & 73.4 & 77.5 & \textbf{78.5} & 63.6 & 68.5 \\
NanoBEIR NanoTouche2020 & 41.6 & 34.2 & 25.7 & 33.6 & 50.3 & 46.3 & \textbf{52.0} & 50.1 & \textbf{52.0} & 42.8 & 45.4 \\
\bottomrule
\end{tabularx}
\end{table}

\clearpage

\section{Additional Implementation Details}
The hindsight teacher, student, and InfoNCE baseline share the same training mixture, sampling, preprocessing, optimization, and query/candidate formatting.

\paragraphnew{Training data and preprocessing.}
\Cref{tab:supp_mixture} lists the sampling weights for MMEB-train~\cite{jiang2024vlm2vec}, ViDoRe~\cite{faysse2024colpali}, VisRAG~\cite{yu2024visrag}, and LLaVA-Hound~\cite{zhang2024direct}.
At each step, we draw one source $s$ with probability $w_s/\sum_{s'}w_{s'}$ and construct the global batch from that source.
We pool four classification datasets to reduce false negatives from repeated class labels within a batch.
Images retain their original aspect ratios, with 4 to 1,280 visual tokens per image ($28\times28$ pixels per token).
For each video, we sample eight frames uniformly with a total budget of 1,228,800 pixels.

\begin{table}[H]
\centering
\caption{\textbf{Training sources and sampling weights.} $^\dagger$The pooled classification source comprises N24News, HatefulMemes, VOC2007, and SUN397.}
\label{tab:supp_mixture}
\small
\renewcommand{\arraystretch}{1.12}
\begin{tabular*}{\linewidth}{@{\extracolsep{\fill}}llr@{\hspace{2em}}llr@{}}
\toprule
Modality & Source & Weight & Modality & Source & Weight \\
\midrule
Image & ImageNet-1K              & 1.0 & Image & VisDial             & 1.0 \\
Image & Classification$^\dagger$ & 1.0 & Image & CIRR                & 0.5 \\
Image & OK-VQA                   & 0.5 & Image & VisualNews t2i      & 1.0 \\
Image & A-OKVQA                  & 0.5 & Image & VisualNews i2t      & 1.0 \\
Image & DocVQA                   & 1.0 & Image & MSCOCO t2i          & 1.0 \\
Image & InfographicsVQA          & 0.5 & Image & MSCOCO i2t          & 1.0 \\
Image & ChartQA                  & 0.5 & Image & NIGHTS              & 0.5 \\
Image & Visual7W                 & 1.0 & Image & WebQA               & 0.5 \\
Image & MSCOCO grounding         & 1.0 &       &                     &     \\
\midrule
VisDoc & ViDoRe         & 10.0 & VisDoc & VisRAG        & 12.0 \\
\midrule
Video & LLaVA-Hound, video$\to$caption & 5.0 &
Video & LLaVA-Hound, caption$\to$video & 5.0 \\
Video & LLaVA-Hound, video QA          & 5.0 &
      &                                &     \\
\bottomrule
\end{tabular*}
\end{table}

\paragraphnew{Optimization.}
We initialize the teacher and student from the same pretrained Qwen2-VL backbone~\cite{wang2024qwen2vl} and train each for one epoch with a global batch size of 512. We freeze the vision encoder and update the multimodal projector and language decoder. We use AdamW with weight decay 0.01 and learning rates of $10^{-5}$ for 2B and $5\times10^{-6}$ for 7B. The learning rate follows a cosine schedule after 100 linear warmup steps.
The teacher and student stages take approximately 110 and 90 H200 GPU-hours for HEME-2B, and 310 and 190 GPU-hours for HEME-7B.

\paragraphnew{Input formatting and readout.}
We format each input as a single user turn with the Qwen2-VL chat template and omit the system prompt.
The final newline serves as the \texttt{[Embed]} position. We replace its input embedding with a learnable vector initialized from the newline embedding and $\ell_2$-normalize its last-layer hidden state.
In the hindsight pass, the attention mask blocks both \texttt{[Embed]} and the template tokens following the query content from attending to the candidate prefix.

\paragraphnew{Hindsight implementation.}
We compute M-RoPE~\cite{wang2024qwen2vl} position indices for the concatenated sequence $[C_i;Q_i]$. The cached candidate keys keep their prefix positions, and the query positions continue from the end of the prefix.
For $\mathcal{L}_{\mathrm{align}}$, we pair each query content token with the same token in the other pass and propagate alignment gradients through both passes.

\end{document}